\pdfoutput=1
\documentclass[runningheads]{llncs}

\usepackage{eccv}

\usepackage{eccvabbrv}
\usepackage[accsupp]{axessibility} 
\usepackage{graphicx}
\usepackage{booktabs}
\usepackage{etoc}

\usepackage[dvipsnames]{xcolor}
\usepackage{amsmath}
\usepackage{amssymb}
\usepackage{nicefrac}
\usepackage{float}
\usepackage{tabularx}
\usepackage{pifont}
\usepackage{subfiles}
\usepackage{multirow}
\usepackage{caption}

\usepackage{hyperref}

\usepackage{orcidlink}

\newcommand{\cmark}{\ding{51}}%
\newcommand{\xmark}{\ding{55}}%

\newcommand{\boldstart}[1]{\vspace{0.08in}\noindent{\bf #1}}

\begin{document}

\addtocontents{toc}{\protect\setcounter{tocdepth}{0}}

\title{Boundary Attention: Learning curves, corners, junctions and grouping}

\author{Mia Gaia Polansky$^{1}$\thanks{Much of this work was done while the author was a student researcher at Google.} \quad Charles Herrmann$^2$
\quad  Junhwa Hur$^2$ \\
\quad  Deqing Sun$^2$
\quad  Dor Verbin$^2$ \quad  Todd Zickler$^2$}
\authorrunning{M. Polansky et al.}
\institute{$^1$Harvard University \hspace{1cm}
\quad 
$^2$Google}

\maketitle

\begin{abstract}
We present a lightweight network that infers grouping and boundaries, including curves, corners and junctions. It operates in a bottom-up fashion, analogous to classical methods for sub-pixel edge localization and edge-linking, but with a higher-dimensional representation of local boundary structure, and notions of local scale and spatial consistency that are learned instead of designed. Our network uses a mechanism that we call boundary attention: a geometry-aware local attention operation that, when applied densely and repeatedly, progressively refines a pixel-resolution field of variables that specify the boundary structure in every overlapping patch within an image. Unlike many edge detectors that produce rasterized binary edge maps, our model provides a rich, unrasterized representation of the geometric structure in every local region.  We find that its intentional geometric bias allows it to be trained on simple synthetic shapes and then generalize to extracting boundaries from noisy low-light photographs.
\end{abstract}

\section{Introduction}
\label{sec:intro}

Converting a precise contour that is defined mathematically in continuous 2D space to a discrete pixel representation is a common task in computer graphics, and there are established tools for rasterization \cite{foley1982fundamentals,pineda1988parallel}, anti-aliasing \cite{heckbert1989fundamentals,catmull1974subdivision} and so on. However, the inverse problem in computer vision of robustly inferring precise, unrasterized contours from discrete images remains an open challenge, especially in the presence of noise. 

Earlier work by Canny~\cite{canny1986computational} and many others~\cite{concetta1988feature,parent1989trace,freeman1991design,freeman1992steerable,martin2004learning} explored the detection of unrasterized parametric edge models, and there are a variety of bottom-up algorithms that try to connect them by encouraging geometric consistency via edge-linking or message-passing. But since the dawn of deep learning, boundaries have almost exclusively been represented using discrete, rasterized maps; and spatial-consistency mechanisms that were previously based on explicit curve geometry have largely been replaced by black-box neural networks. 

In this paper, we take inspiration from early computer vision work and revisit the task of finding unrasterized boundaries via bottom-up geometric processing. But unlike early work, we leverage self-attention to learn these processes instead of hand-crafting them, thereby combining the benefits of geometric modeling with the efficiency and representational power of deep learning.

\begin{figure*}[t]
\centering
\includegraphics[width=\textwidth]{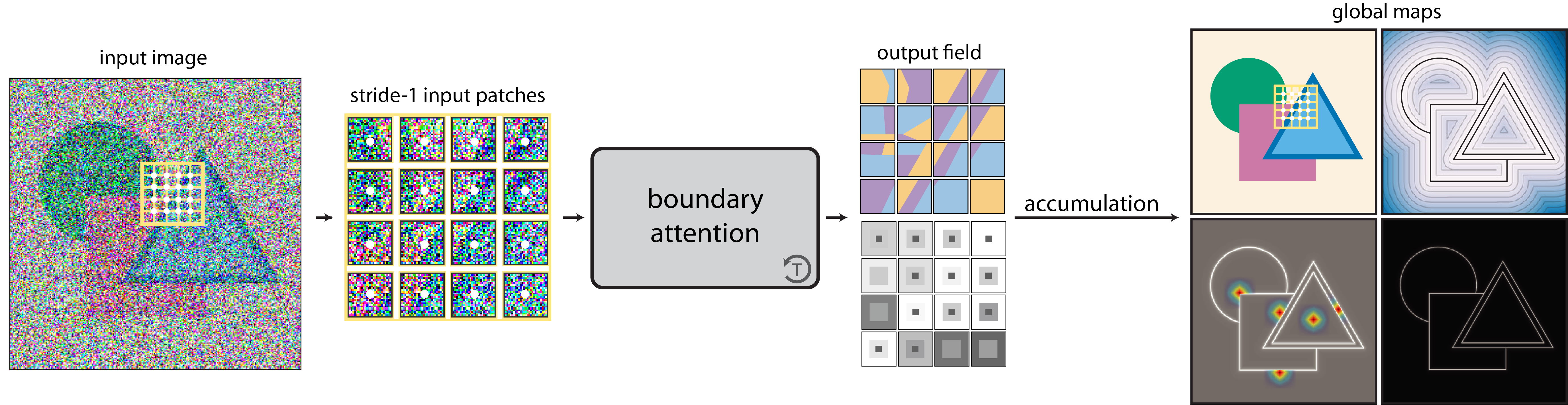}
\caption{\label{fig:overview} Pipeline overview. The image unfolds into stride-1 patches, and boundary attention operates iteratively on their embeddings to produce for each patch: (\emph{i}) a parametric three-way partitioning, and (\emph{ii}) a parametric windowing function that defines its effective patch size. (Figure~\ref{fig:junction-space} shows parameterization details.) This output field implies a variety of global maps, shown in clockwise order: a boundary-aware smoothing of the input colors; an unsigned boundary-distance map; a boundary map; and a map of spatial affinities between any query point and its neighbors.}
\end{figure*}

We focus on the low-level task of finding boundaries that separate regions of uniform color, as depicted by the toy problem in Figure~\ref{fig:overview}. This task becomes difficult at high noise levels, especially within local patches, and we address it by creating a model that can learn to exploit a wide range of low-level cues, such as curves being predominantly smooth with low-curvature; curves tending to meet at corner and junction points; contours tending to have consistent contrast polarity throughout their extent; and colors tending to vary smoothly at locations away from boundaries.

The core of our model is a mechanism we call boundary attention. It is a boundary-aware local attention operation that, when applied densely and repeatedly, progressively refines a field of variables that specify the local boundaries within dense (stride-1) patches of an image. The model's output is a dense field of unrasterized geometric primitives that, as depicted in the right of Figure~\ref{fig:overview}, can be used in a variety of ways, including to produce an unsigned distance function and binarized map for the image boundaries, a boundary-aware smoothing of the input image colors, and a map of spatial affinities between any query point and the pixels that surround it.

\begin{figure}[t]
\centering
\includegraphics[width=.7\columnwidth]{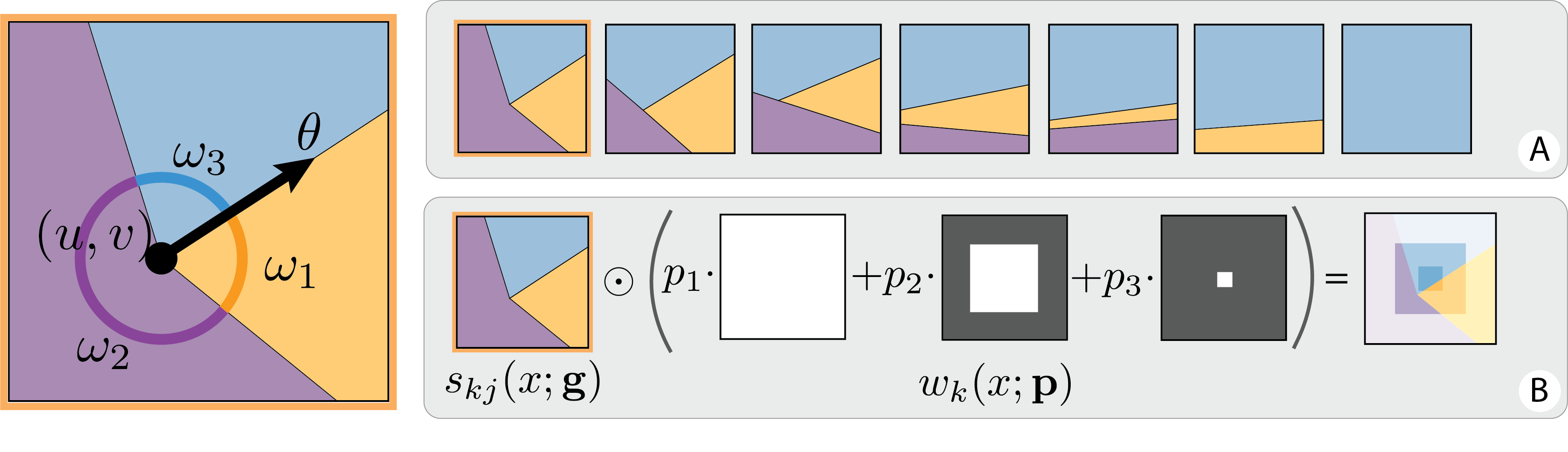}
\caption{\label{fig:junction-space} Parameterization details. \emph{Left}: Each patch $k$ is associated with an unrasterized three-way partitioning of its area (colored blue, orange and purple here). The partitioning parameters comprise a vertex $(u,v)$, orientation $\theta$, and angles $(\omega_1,\omega_2,\omega_3)$, defined up to scale. \emph{A}: A walk through junction space by linearly interpolating between junctions is spatially smooth, and can represent edges, bars, corners, Y-junctions, T-junctions and uniform regions. \emph{B}: Each junction is modulated through a learned windowing function. The windowing parameters $\mathbf{p}=(p_1,p_2,p_3)$ are convex weights over a dictionary of binary pillboxes.
}
\end{figure}

An important feature of our model is that its output patch primitives (see Figure~\ref{fig:junction-space}) have enough flexibility to represent a wide range of local boundary patterns and scales, including thin bars, edges, corners and junctions, all with various sizes, and all without rasterization and so with unlimited resolution. This gives our model enough capacity to learn to localize boundaries with high precision, and to regularize them in the presence of noise without erroneously rounding their corners or missing their junction points.

We intentionally design our model with components that are local and invariant to discrete spatial shifts, enabling it to be trained on small-sized images and then deployed on much larger and differently-shaped ones. By tying the weights across multiple portions of the network, we increase its resilience to noise, which we validate through ablations. Our resulting model is very compact, comprising only 207k parameters, and it runs much faster than comparable optimization-based approaches. Further, our model can be trained to a useful state with very simple synthetic data, made up of random circles and triangles that are uniformly colored and then corrupted by noise. Despite the simplicity of this training data, we find that the model's learned internal activations exhibit intuitive behavior, and that the model generalizes to real world photographs, including those taken in low-light and have substantial shot noise.

Our main contributions can be summarized as follows:

\begin{enumerate}
    \item We introduce a bottom-up, feedforward network that decomposes an image into an underlying field of local geometric primitives that explicitly identify curves, corners, junctions and local grouping. 
    \item We do this by introducing a new parameterization for local geometric primitives, and a new self-attention mechanism that we call boundary attention.
    \item We identify the best architecture among a family of candidates, and we show that it generalizes from simple synthetic images to real low-light photographs.
\end{enumerate}

\section{Related Work}
\label{sec:related-work}

Early approaches to edge detection rely primarily on low-level or local information. Canny~\cite{canny1986computational} and others (\eg,~\cite{concetta1988feature,parent1989trace,freeman1991design,freeman1992steerable}) use  carefully chosen filters, and later ones like pB~\cite{martin2004learning} and gPb~\cite{4587420} combine such filters using a handful of trainable parameters. These methods provide local estimates of straight-edge position and orientation and they can subsequently be filtered and joined using geometry-based processes such as edge-linking~\cite{canny1986computational}. However, they often struggle near corners and junctions due to the difficulty of designing filters for these more complicated structures. Structured edges~\cite{dollar2014fast} and other early learning-based methods~\cite{dollar2006supervised, lim2013sketch} are able to detect more-complicated structures, but they represent their local outputs using discrete, rasterized boundary maps that are not directly compatible with continuous boundary parameterizations or geometry-based linking. 

Recently, the field of junctions (FoJ)~\cite{verbin2021field} introduced a way to detect more-complicated local structures while also representing them parametrically, without rasterization. By using its unrasterized representations in a customized non-convex optimization routine, the method can detect curves, corners and junctions with high precision and with unprecedented resilience to noise. Our model is strongly inspired by this work, and it improves upon it by introducing an enhanced family of parameterized geometric primitives (Figure~\ref{fig:junction-space}) and by replacing its hand-designed objective and associated non-convex optimization routine with a learned model that is fast, feed-forward and differentiable. We also tackle the challenge of representing images that vary in scale across different areas: whereas FoJ requires setting a single patch size for the entire image, our model learns to choose patch sizes adaptively in response to the input image.

We emphasize that our work is very different from a recent trend in computer vision of using deep networks to detect boundaries that are semantically meaningful (\eg,~\cite{su2021pixel, pu2022edter, zhou2023treasure}). These models aim to identify boundaries between semantic concepts, such as people and animals, while ignoring boundaries that are inside of these concepts, such as stripes on a shirt. In contrast to this trend, we follow earlier work by focusing entirely on identifying boundaries from the bottom up, using low-level cues alone. Since our model is trained using simple synthetic data, it does not exploit object familiarity or learn to ignore intra-object contrast. This approach has advantages: It is not specialized to any predetermined set of semantic concepts or tasks, and instead of producing a pixelized boundary map, it produces a field of unrasterized primitives that provide better precision as well as richer information about local grouping. We leave for future work the exploration of how to combine our bottom-up model with other cues that are top-down and semantic.

\section{Representation}
\label{sec:representation}

Our system is depicted in Figure~\ref{fig:overview}. It uses neighborhood cross-attention, a patch-wise variant of cross-attention, with $D$-dimensional, stride-1  embeddings. Critically, each $D$-dimensional embedding explicitly encodes a tuple of values that specifies the geometric structure and spatial extent of the unrasterized local boundaries within a patch. 

Our model relies on a  learned (linear) mapping from the embedding dimension $D$ to a hand-crafted, lower-dimensional space of unrasterized boundary patterns that we call \emph{junction space}. Junction space has the benefit of specifying per-patch boundary patterns without rasterization and thus with unlimited spatial precision. As depicted in Figure~\ref{fig:junction-space} and described in~\cite{verbin2021field}, it also has the benefit of including a large family of per-patch boundary patterns, including uniformity (\ie, absence of boundaries), edges, bars, corners, T-junctions and Y-junctions. Different from~\cite{verbin2021field}, we additionally  modulate the spatial extent of each patch by learning an associated windowing function, to be described in Section~\ref{sec:primitives}.

To enable communication between patch embeddings, each of which corresponds to a local patch, we leverage the idea that each image point is covered by multiple patches, and that overlapping patches must agree in their regions of overlap. This has the effect of tying neighboring patches together, analogous to cliques in a Markov random field. 

We refer to the core mechanism of our model as \emph{boundary attention}. We introduce it by first defining our hand-crafted parameterization of junction space and some associated operators. Then Section~\ref{sec:network} describes the architecture that we use to transform an image into a coherent junction-space representation.

\subsection{Boundary Primitives}
\label{sec:primitives}

We use parentheses $(x)$ for continuous signals defined on the 2D image plane $[0,W]\times[0,H]$ and square brackets $[n]$ for discrete signals defined on the pixel grid. We use $c[n]$ for the coordinates of the pixel with integer index $n$.

Denote the $C$-channel input image by $\{\mathbf{f}[n]\}$, where $\mathbf{f}[n]\in\mathbb{Q}^C$ is the vector image value at the discrete pixel grid index $n$. Our approach is to treat the image as a field of dense, stride-1 overlapping local patches. We use $\Omega_{k}(x)$ to denote the spatial support of the patch that is centered at the pixel whose integer index is $k$.

There are many ways to partition a local patch $\Omega_k(x)$, and one can define parametric families of partitions. For example the set of oriented lines provides a two-parameter family of partitions, with each member of the family separating the region into points that lie on one side of a line or the other. This family of partitions would be appropriate for describing edges. Here we define a larger family of partitions that encompasses a greater variety of local boundary structures.

As depicted in the right of Figure~\ref{fig:junction-space}, our partitions are parameterized by $\mathbf{g}\in\mathbb{R}^2\times\mathbb{S}^1\times\triangle^{2}$, where $\mathbb{S}^1$ is the unit circle and $\triangle^{2}$ is the standard $2$-simplex. We use the notation $\mathbf{g}=(\boldsymbol{u}, \theta, \boldsymbol{\omega})$, where $\boldsymbol{u}=(u,v)\in \mathbb{R}^2$ is the \emph{vertex}, $\theta\in\mathbb{S}^1$ is the \emph{orientation}, and   $\boldsymbol{\omega}=(\omega_1,\omega_2,\omega_3)$ are barycentric coordinates (defined up to scale) for the three relative angles, ordered clockwise starting from $\theta$. Our convention is to express the vertex coordinates relative to the center of region $\Omega_k(x)$, which is located at $c[k]$, and we note that the vertex is free to move outside of this region. We also note that up to two angles $\omega_j$ can be zero. This all makes it possible to smoothly represent a variety of partition types, including edges, bars, corners, 3-junctions and uniformity (\ie, trivial or singleton partitions), and do so with unlimited spatial resolution. 

\begin{figure}[t]
\center
\includegraphics[width=\columnwidth]{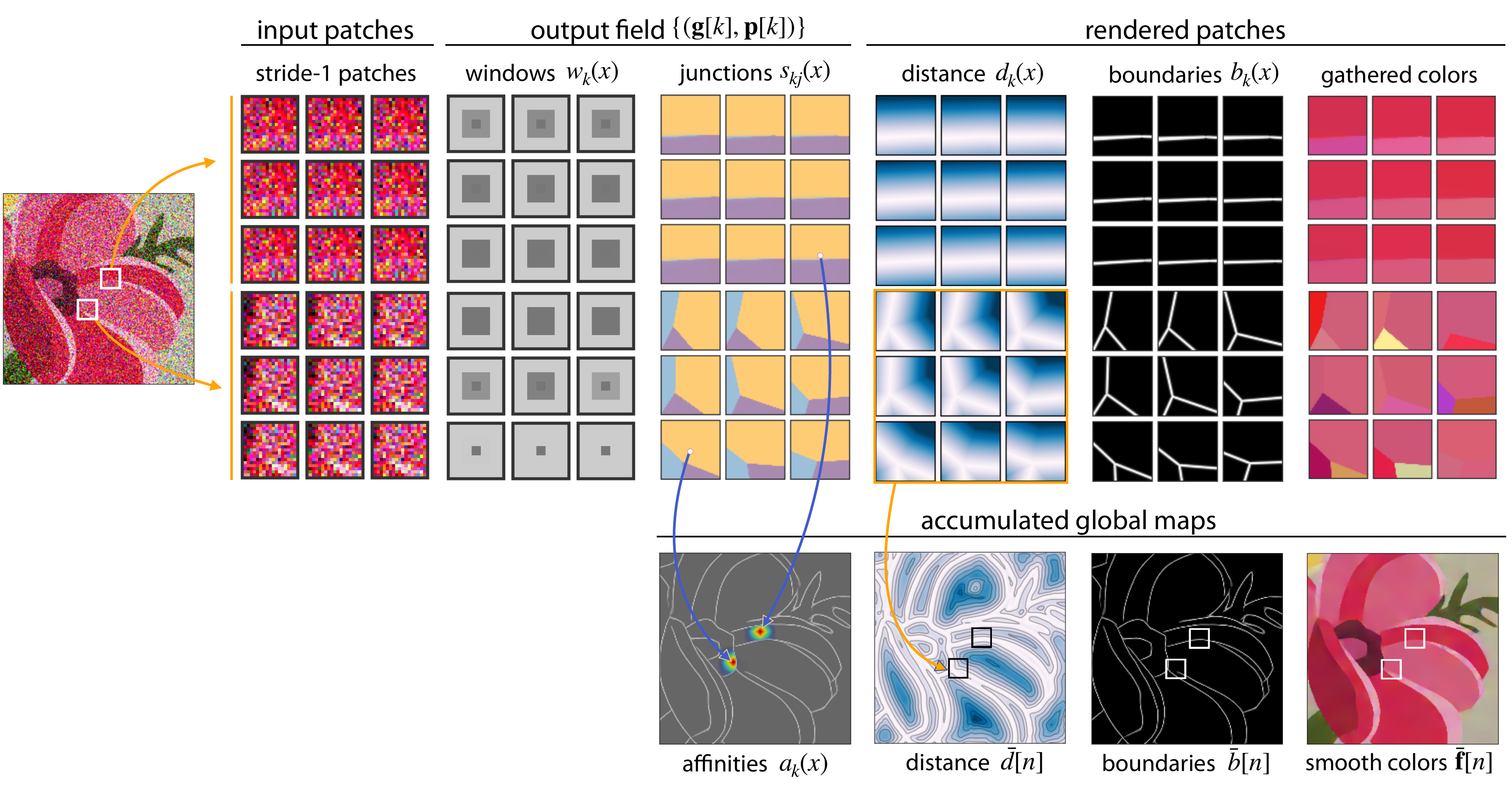}
\caption{\label{fig:outputs} Example of our model's output, with examples from two different regions. \emph{Top row}: Some of each region's overlapping input patches, their corresponding outputs (visualized in the style of Figure~\ref{fig:junction-space}), and three types of per-patch attributes that the outputs imply: unsigned distance; boundaries; and gathered wedge colors. \emph{Bottom row}: Four types of global maps that are implied by accumulating values from the output field and rendered patches.}
\end{figure}

Fixing a value for $\mathbf{g}$ induces three wedge support functions, denoted by
\begin{equation}
s_{kj}(x;\mathbf{g})\in\{0,1\},\ j=1, 2, 3.
\end{equation}
These evaluate to $1$ for points $x$ that are in $\Omega_k(x)$ and in the $j$th wedge defined by $\mathbf{g}$; and $0$ otherwise. It also induces an unsigned distance function, denoted by
\begin{equation}
d_{k}(x;\mathbf{g})\ge 0,
\end{equation}
which represents the Euclidean distance from point $x$ to the nearest point in the boundary set defined by $\mathbf{g}$. Figure~\ref{fig:junction-space} uses three colors to visualise the wedge supports $s_{kj}$ of a junction $\mathbf{g}$, and Figure~\ref{fig:outputs} shows the unsigned distance functions for $3\times 3$ grids of junction parameters. Analytic expressions for them are included in the supplement.

In order to enable the size of each region $\Omega_k$ to adapt to the local geometry and noise conditions, we equip each one with a parameterized local windowing function $w_k(x; \mathbf{p})\in[0,1]$, with parameters $\mathbf{p}\in{\cal P}=\triangle^{W-1}$ that are the coefficients of a convex combination of $W$ square pillbox functions. That is,
\begin{equation} \label{eq:windowfnconcrete}
    w_k(x;\mathbf{p}) = \sum_{i=1}^W p_i \mathbf{1}[\|x - c[k]\|_\infty \leq D_i],
\end{equation}
where $\|\cdot\|_{\infty}$ is the $\ell^{\infty}$-norm, and $\mathbf{1}[\cdot]$ is the indicator function that returns $1$ if the argument is true; and $0$ otherwise. In our experiments we use $W=3$ and diameters $\mathbf{D} = (3, 9, 17)$. \Cref{fig:outputs} shows some examples.

\subsection{Gather and Slice Operators}

\begin{figure}[t]
\centering
\includegraphics[width=\columnwidth]{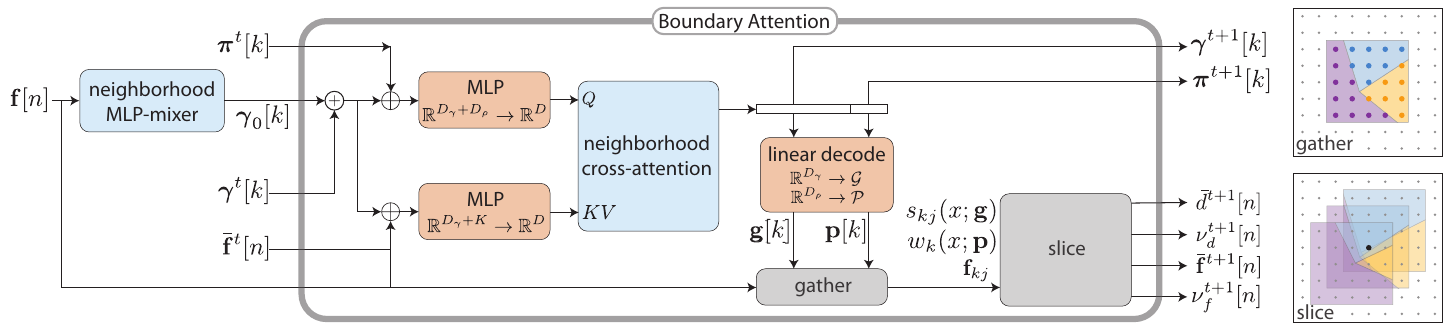}\caption{\label{fig:network} Model Architecture. All blocks are invariant to discrete spatial shifts, and only colored blocks are learned. Orange blocks operate at individual locations $n$, while blue ones operate on small spatial neighborhoods. Symbol $\oplus$ is concatenation, and gather and slice operators (Eqs.~\ref{eq:gather}--\ref{eq:globalf}) are depicted at right. The first iteration uses $\boldsymbol{\gamma}^0[n]=\boldsymbol{\gamma}_0[n]$, $\bar{\mathbf{f}}^0[n]=\mathbf{f}[n]$, and $\boldsymbol{\pi}^0[n]=\boldsymbol{\pi}_o$ with $\boldsymbol{\pi}_o$ learned across the training set. Boundary attention repeats $T=8$ times, with one set of weights for the first four iterations and a separate set of weights for the last four iterations, resulting in 207k trainable parameters total.}
\end{figure}

Our network operates by refining the field $\{(\mathbf{g}^t[k], \mathbf{p}^t[k])\}$ over a fixed sequence of steps $t=1,...,T$. It uses two operators that we define here and depict in the right of Figure~\ref{fig:network} to facilitate intra-patch communication between pixels within a single patch, and inter-patch communication across different patches representing overlapping image regions. The first operator is a patch-wise \emph{gather} operator, in which each wedge of each patch computes the weighted average of the image values it contains (recall that $c[n]$ are the $n$th pixel's coordinates, and wedges are indexed by $j$):
\begin{equation}
    \mathbf{f}_{kj}=\frac{\sum_n \mathbf{f}[n]w_{k}(c[n];\mathbf{p}[k])s_{kj}(c[n];\mathbf{g}[k])}{\sum_n w_{k}(c[n];\mathbf{p}[k])s_{kj}(c[n];\mathbf{g}[k])}.\label{eq:gather}
\end{equation}

The second operation is a global pixel-wise \emph{slice} operation, where each pixel computes the means and variances, over all regions that contain it, of the  per-region distance maps $d_k(x;\mathbf{g}[k])$ and gathered wedge features $\mathbf{f}_{kj}$. The expressions for the means are:

\begin{align}
    \bar{d}[n] &=\frac{\sum_{k} w_{k}(c[n];\mathbf{p}[k])d_{k}(c[n];\mathbf{g}[k])}{\sum_{k} w_{k}(c[n];\mathbf{p}[k])}, \label{eq:slice-d}\\
    \bar{\mathbf{f}}[n] & =\frac{\sum_{k} w_{k}(c[n];\mathbf{p}[k])\sum_j \mathbf{f}_{kj}s_{kj}(c[n];\mathbf{g}[k])}{\sum_{k} w_{k}(c[n];\mathbf{p}[k])\sum_j s_{kj}(c[n];\mathbf{g}[k])}.\label{eq:globalf}
\end{align}
Here, the sums are over $\{k \mid \Omega_{k}\ni c[n]\}$ so that only patches that contain $c[n]$ contribute to the sum. Similar expressions for pixel-wise distance map variance $\nu_d[n]$
and feature variance $\nu_f[n]$, which is computed across patches containing $n$ and across their $K$ channels, are included in the supplement. Intuitively, slicing represents an accumulation of regional information into a pixel, as dictated by the partitions of all of the patches that contain the pixel.

\subsection{Visualizing Output}
\label{sec:output}

Our network's output is a field of tuples representing the junction and windowing parameters $\{(\mathbf{g}[k], \mathbf{p}[k]\}$ for all stride-1 patches of the input image. We  visualize them in Figure~\ref{fig:outputs} by rasterizing the continuous windowing functions $w_k(x;\mathbf{p})$ (second column) and binary wedge supports $s_{kj}(x;\mathbf{g})$, which are colored purple, orange, and blue (third column). To the left, we show the input image and the regions from which the patches were extracted.

Additionally, we can use the output junction parameters to rasterize unsigned distance patches $d_k(x;\mathbf{g})$ (fourth column), boundary patches $b_k(x;\mathbf{g})$ (fifth column), and wedge supports  $s_{kj}(x;\mathbf{g})$ that are re-colored with their respective wedge features $\mathbf{f}_{kj}$ (sixth column). Note that all of these are defined continuously, so can be rasterized at any resolution.

We expect the shapes of junction boundaries in overlapping regions to agree, so that the variances $\nu_d[n], \nu_f[n]$ are small at every pixel. Then, the fields of means $\{\bar{d}[n]\}, \{\bar{\mathbf{f}}[n]\}$ can be interpreted, respectively, as a global unsigned distance map for the image boundaries  (bottom row, fourth column) and a boundary-aware smoothing (bottom row, last column) of its input channel values. 
Figure~\ref{fig:outputs} shows an example (bottom row, fifth column), where we visualize the zero-set of the global unsigned distance map---we call this the global boundaries---by gathering boundary patches $b_k$ defined as:
\begin{equation} \label{eq:nonlinearboundary}
    b_k(x;\mathbf{g})=\left(1 + (d_k(x;\mathbf{g})/\eta)^{2}\right)^{-1},
\end{equation}
setting $\eta=0.3$.

For any query pixel $k$, we can also probe the containing wedge supports $\{s_{kj}(\cdot;\mathbf{g}[k])\}$ and windowing functions $\{w_k(\cdot, \mathbf{p}[k])\}$ to compute a spatial affinity map $a_{k}(x)$ that surrounds the query pixel. This represents the affinity between point $c[k]$ and a surrounding neighborhood with diameter that is twice that of $\Omega(x)$. It is also the boundary-aware spatial kernel that turns the neighborhood of input features $\{\mathbf{f}[\cdot]\}$ into the gathered value $\bar{\mathbf{f}}[n]$ via \begin{equation} \label{eq:attentionmap}
\bar{\mathbf{f}}[n] = \sum_{k} a_n(c[k]) \mathbf{f}[k].
\end{equation}
The expression for $a_n(x)$ follows from inserting  Equation~\ref{eq:gather} into~\ref{eq:globalf}. The spatial affinity maps for two probed points are shown in the leftmost image of the bottom row of Figure~\ref{fig:outputs}. Like slicing, querying the affinity map at a pixel is a form of regional accumulation from patches to a pixel.

\section{Network Architecture}
\label{sec:network}

We seek a differentiable architecture that can effectively initialize and then refine the fields $\{(\mathbf{g}^t[k],\mathbf{p}^t[k])\}$ using a manageable number of iterations. This approach is motivated by the edge localization and edge-linking steps of early edge detectors like Canny's, but has a more sophisticated local boundary model, and mechanisms for spatial consistency that are learned instead of designed. It is also analogous to the original field of junctions algorithm~\cite{verbin2021field}, which uses coordinate descent for initialization and iterations of gradient descent for refinement. We want to replace both steps with something that is differentiable, faster, and able to scale to larger images.

After considering a variety of alternatives, we settled on a particular family of architectures based on dot-product self-attention; and within this family, we performed an extensive set of ablations to determine the best performing model. We describe our final model here, and we provide the ablation details in Section~\ref{sec:ablations}. Importantly, all of our model's components are invariant to discrete spatial shifts of the image, operating either on individual locations $k$, or on small neighborhoods of locations with spatially-shared parameters. This means that our model can be trained on small images and then deployed on much larger ones. Also, our model has only $207,000$ learnable parameters, making it orders of magnitude smaller than most deep semantic boundary detectors. As a point of reference, Diffusion Edge, a recent diffusion-based semantic boundary detection model uses on the order of 300 million parameters~\cite{ye2024diffusionedgediffusionprobabilisticmodel}.

Our model is depicted in Figure~\ref{fig:network}. It represents the field elements as higher-dimensional embeddings $\boldsymbol{\gamma}^t[k]\in\mathbb{R}^{D_\gamma}$ and $\boldsymbol{\pi}^t[k]\in\mathbb{R}^{D_\pi}$, which can be decoded at any iteration using the learned linear mappings $\boldsymbol{\gamma}\mapsto \mathbf{g}$ and $\boldsymbol{\pi}\mapsto \mathbf{p}$. Our final model uses $D_\gamma = 64$ and $D_\pi = 8$ which we show in our experiments provides enough capacity to learn a smooth latent representation of junction space. Using separate embeddings for the junction and windowing fields provides a disentangled representation of both.

Given an input image, the network first applies a ``neighborhood MLP-mixer'', which is a modified MLP-Mixer~\cite{tolstikhin2021mlpmixer} that replaces the global spatial operations with convolutions of kernel size $3$. The other change is that we map the input pixels to the hidden state size with a pixel-wise linear mapping rather than taking patches of the input. This block transforms the input image into a pixel-resolution map with ${D_\gamma}$ channels. We denote this by $\gamma_0[n]$ and refer to it as the initial ``hidden state''. This hidden state is then refined by a sequence of eight boundary attention iterations, which we describe next. (See our experiments for a visualization of the decoded hidden states as they evolve.)

The eight iterations of refinement are broken into two blocks, each with learned weights. In each iteration, we first add a linear mapping of the initial hidden state to the current hidden state, which acts as a skip connection. Next, we clone our hidden state, concatenating a dimension 8 learned windowing embedding to one of the copies and the input image plus the current estimate of the smoothed global features to the other. We treat the copies as the inputs to neighborhood cross-attention: each pixel in the first copy does two iterations of cross attention with a size 11 patch of the second copy. We add a learned $11 \times 11$ positional encoding to the patch, which allows our network to access relative positioning, even though global position cues are absent. We follow each self attention layer with a small MLP.

To transform our output or intermediary hidden state into junction space and render patch visualizations, we use a simple linear mapping. We separate the windowing embedding (the last 8 dimensions) from the junction embedding (the first 64 dimensions) and project each through a linear layer. We map the junction embedding to 7 numbers that represent $\mathbf{g}=(\boldsymbol{u}, \sin(\theta), \cos(\theta), \boldsymbol{\omega})$. These serve as the inputs to our gather and slice operators. 

To extract $\mathbf{p}$ from the windowing embedding, we linearly project the windowing embedding to a length 3 vector, which we use as coefficients in a weighted sum of three square pillbox functions with widths $3$, $9$, and $17$. This implementation of the windowing function ensures spatial overlap between neighboring patches by limiting the minimum patch size to $3$. 

\subsection{Training}
\label{sec:training}
Estimating the best junctions for noisy image patches is a non-convex optimization that is prone to getting stuck in local minima~\cite{verbin2021field}. We find that training our network in three stages with increasingly complex synthetic training data produces the best model. First we train the network on small $21 \times 21$ images containing a single junction corrupted by low amounts of Gaussian noise. Upon convergence, we retrain the network on larger $100 \times 100$ images containing a single circle and triangle corrupted by moderate Gaussian noise. Finally, we retrain the network on a high-noise synthetic dataset containing $125 \times 125$ images consisting of many overlapping triangles and circles. (See supplement for examples.)

We calculate our loss using the outputs of the two final iterations of our network, where the loss of the final output is weighted three times as much as the loss applied to the output before it. This encourages the network to allocate capacity to producing high quality outputs, while still providing supervision to intermediate junction estimates.

We train our model using a combination of four global losses applied to global (\ie sliced) fields, and two patch-wise losses applied to individual patches. The first two losses are supervision losses penalizing mismatches between our network's predictions and the ground truth feature and distance maps:
\begin{align}
    \mathcal{L}_f &= \sum_{n} \alpha[n] \|\bar{\mathbf{f}}[n] - \mathbf{f}_{\text{GT}}[n]\|^2, \label{eq:floss}\\
    \mathcal{L}_d &= \sum_{n} \alpha[n] \left(\bar{d}[n] - d_{\text{GT}}[n]\right)^2,
\end{align}
where $\mathbf{f}_{\text{GT}}$ and $d_{\text{GT}}$ are the ground truth features and distance maps, respectively, and $\alpha[n]$ is a pixel importance function defined as
\begin{equation} \label{eq:alpha}
    \alpha[n] = e^{-\beta \cdot (d_{\text{GT}}[n] + \delta)} + C,
\end{equation}
with $\beta$ and $C$ controlling how much weight to give pixels near boundaries. We set $\beta = 0.1$, $\delta = 1$, and $C=0.3$. (The supplement contains additional tests with a more involved importance mask.)
Using noiseless feature maps for supervision in Equation~\ref{eq:floss} has the effect of encouraging windowing functions to be as large as possible, because larger regions imply averaging over a greater number of noisy input pixel values.

On top of the two supervision losses we apply two consistency losses borrowed from~\cite{verbin2021field}, which minimize the per-pixel variances $\nu_f[n]$ and $\nu_b[n]$. Here, we weight them by $\alpha[n]$ from Equation~\ref{eq:alpha}. These consistency losses encourage the junction shapes $\mathbf{g}$ in overlapping regions to agree. Minimizing $\nu_f[n]$ also provides a second mechanism to encourage windowing functions to be large. Larger windows increase gather area, thereby reducing noise in wedge features $\mathbf{f}_{nj}$ that are sliced to compute variance $\nu_f[n]$ at each pixel $n$.

Finally, we use two patch-wise losses to encourage individual  feature and distance patches to agree with the supervisory ones:
\begin{align}
    \ell_f = \sum_k \chi[k] \sum_{n \in \Omega_k} \alpha[n] \|\bar{\mathbf{f}}[n] - \mathbf{f}_\text{GT}[n]\|^2, \\
    \ell_d = \sum_k \chi[k] \sum_{n\in\Omega_k} \alpha[n] (\bar{d}[n] - d_{\text{GT}}[n])^2,
\end{align}
where $\chi[k]$ is a patch importance function defined as:
\begin{equation}
    \chi[k] = \left(\sum_{n\in\Omega_k}(d_\text{GT}[n] + \delta')\right)^{-1},
\end{equation}
with $\delta' = 1$. These per-patch losses provide a more direct signal for adjusting model weights, compared to per-pixel losses which average over multiple patches. 

\begin{figure}[t]
\center
\includegraphics[width=.9\columnwidth]{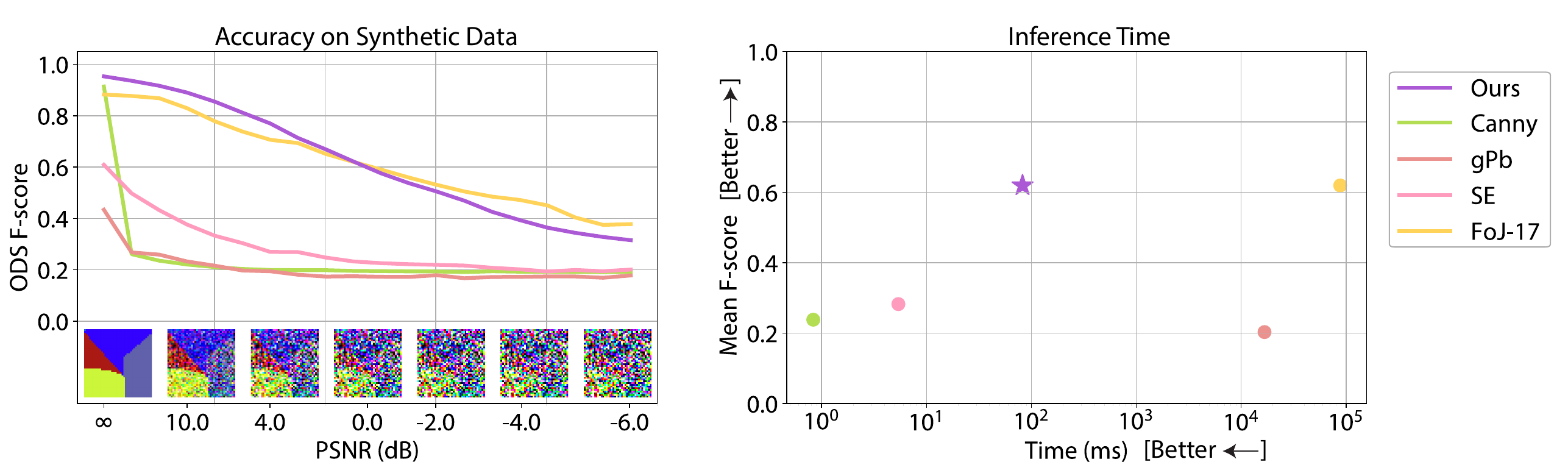}
\caption{\label{fig:syntheticval} \textit{Left:} ODS F-score for our method and multiple baselines at different noise levels computed on noisy synthetic data. The bottom inset show example patches at representative PSNR values. Our method outperforms all baselines at low noise and is better or competitive with other techniques at high noise. \textit{Right:} Comparing the F-score for different techniques with their runtime. Our method has the best average F-score while also being much faster than the second best method Field of Junctions.
}
\end{figure}

\begin{figure}[t]
\center
\includegraphics[width=.9\columnwidth]{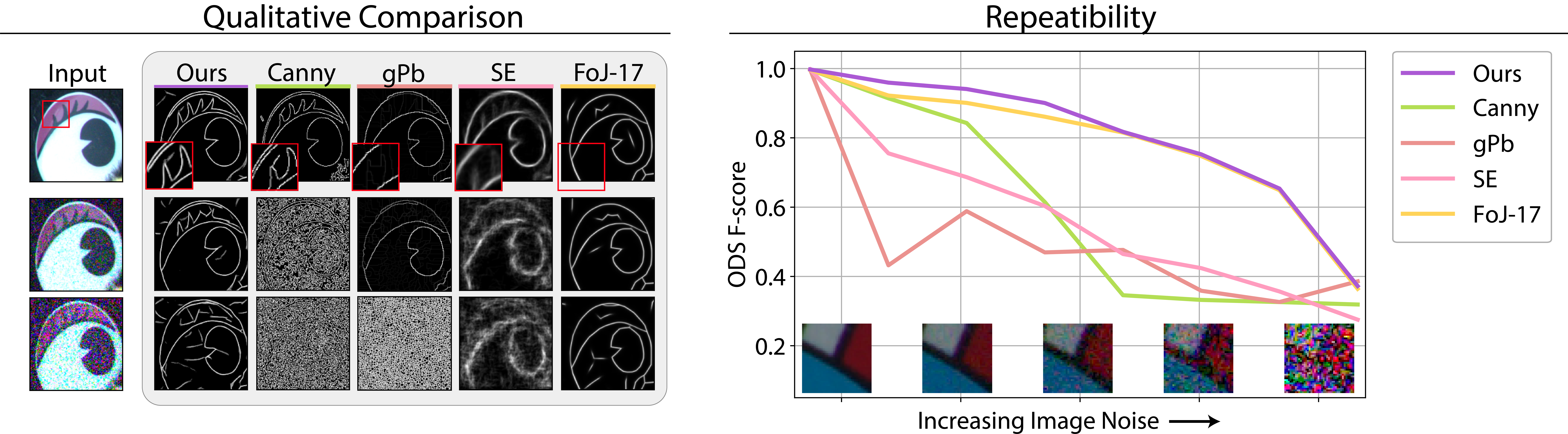}
\caption{\label{fig:ELD-performance} \emph{Left}: Qualitative comparison of our model versus other methods for a noisy crop from the ELD dataset~\cite{wei2020physicsbased}. See the supplement for more examples. Our method provides more detail than other techniques at low noise and is more robust to high levels of noise. \emph{Right}:
Repeatability of the estimated boundaries of crops from the ELD dataset over increasing noise. Over multiple levels of real sensor noise, our method is the most consistent in predicting edges.
}

\end{figure}

\section{Experiments}
\label{sec:experiments}

\vspace{-5pt}
\paragraph{Implementation details.}

In the final stage of training, we use noisy synthetic data of many randomly-colored overlapping triangles and circles. We render $240\times 320$ images containing $15$ to $20$ shapes each, but use $125\times 125$ crops for training. To those crops we add Gaussian or Perlin noise~\cite{perlin1985image}, and with probability $0.1$ we average over the color channels to produce grayscale inputs. Our dataset contains $10^5$ images, $90\%$ of which are used for training, and the rest for validation.

\vspace{-5pt}
\paragraph{Baselines.}

Since we focus on bottom-up edge and corner detection, we compare against other techniques with a similar focus and that, like us, do not train on large semantic datasets: Canny~\cite{canny1986computational}, gPb~\cite{4587420}, Structured Edges (SE)~\cite{dollar2014fast}, and Field of Junctions (FoJ)~\cite{verbin2021field}. 

\vspace{-5pt}
\paragraph{Quantitative results.}

For quantitative evaluation we cannot use semantic edge detection benchmarks like BSDS500~\cite{arbelaez2010contour}, because our model and the comparisons are designed to predict all edges, including those that are not semantically meaningful. We instead rely on synthetic data, where the ground truth edges can be determined with perfect precision, and inputs can be controllably noised. Our evaluation data comprises geometric objects and per-pixel additive Gaussian noise with a random variance.

Figure~\ref{fig:syntheticval} compares the performance and inference time of our method and baselines under different noise levels. The tuneable parameters for Field of Junctions were chosen to maximize its performance on noisy images with $17\times 17$ patches. Notably our method's adaptive windowing function gives it an edge compared to the Field of Junctions at low noise, enabling it to capture finer details, with only slightly worse performance under extreme noise conditions. Our method is also orders of magnitude faster than FoJ, as shown at right.

\paragraph{Qualitative results on real images.}
As shown in Figure~\ref{fig:ELD-performance}, despite being trained on synthetic data, our method can detect edges in real photographs with multiple levels of real sensor noise present in ELD~\cite{wei2020physicsbased}. Our method produces crisp and well-defined boundaries despite high levels of noise. The supplement includes additional examples that show that our method makes reasonable boundary estimates for other captured images.

\paragraph{Repeatability on real images.} We quantify each method's noise resilience on real low-light images by measuring the repeatability of its boundaries over a collection of scenes from the ELD dataset. For each scene, we run the model on the lowest-noise image and then measure, via ODS F-score, how much its predictions change with increasing noise level. Figure~\ref{fig:ELD-performance} shows the averages of these scores across the collection. Our model provides more consistent results for increasingly noisy images than other methods, in addition to capturing fine details that other methods miss.

\begin{figure}[t]
\center
\includegraphics[width=.7\columnwidth]{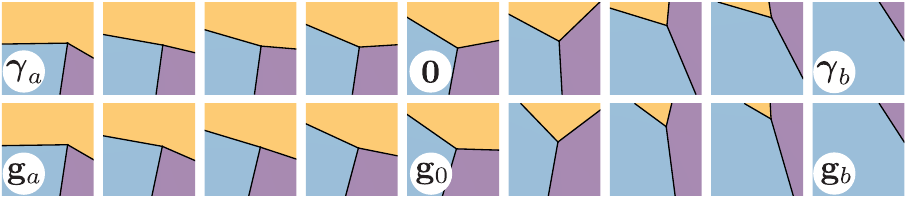}
\caption{\label{fig:interpolation} \emph{Top}: Linear interpolation in our network's learned embedding space $\mathbb{R}^{D_\gamma}$ from value $\boldsymbol{\gamma}_a$ to zero and then to $\boldsymbol{\gamma}_b$. \emph{Bottom}: A geometric interpolation in junction space $\mathbf{g}\in{\cal G}$ that passes through $\mathbf{g}_0=(\mathbf{0},0, \nicefrac{1}{3}\cdot\mathbf{1})$. The embedding has learned to be smooth and have an intuitive zero.}
\end{figure}

\paragraph{Properties of learned embedding.} We find that our model learns a spatially smooth embedding $\boldsymbol{\gamma}\in\mathbb{R}^{D_\gamma}$ of junction space $\mathbf{g}\in {\cal G}$. In Figure~\ref{fig:interpolation} we generate equally-spaced samples $\boldsymbol{\gamma}_i$ by linearly interpolating from a particular $\boldsymbol{\gamma}_a$ to $\boldsymbol{0}$ and then to a particular $\boldsymbol{\gamma}_b$; and then to each sample we apply the learned embedding to compute and visualize the implied junction $\mathbf{g}_i$. We see that the embedding space is smooth, and interestingly, that it learns to associate its zero with nearly-equal angles and a vertex close to the patch center. For visual comparison, we show an analogous geometric interpolation in junction space ${\cal G}$ (see the supplement for expressions) from $\mathbf{g}_a$ to $\mathbf{g}_0\triangleq(\mathbf{0},0,\nicefrac{1}{3}\cdot\mathbf{1})$ and then to $\mathbf{g}_b$.

\begin{figure}[t]
\centering
\includegraphics[width=\textwidth]{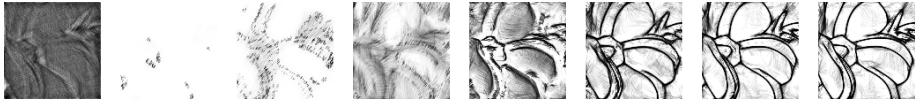}
\caption{\label{fig:evolution} Evolution of boundaries during iterations, in reading order. Early  iterations  are  exploratory  and  unstructured, while later iterations feature consistent per-patch boundaries, resulting in clean average boundary maps.}
\end{figure}

\begin{table}[t]
    \centering
{\scriptsize
\centering
\setlength{\tabcolsep}{4pt}
\begin{tabular}{c c c c | c c c }
NM &BA-1 &BA-2 &Tied & \! Low Noise$\uparrow$  & \! Med-Noise$\uparrow$  & \! High Noise$\uparrow$  \\ \hline
\multicolumn{1}{l}{\cmark} &\xmark &\xmark & N/A & $0.355$   & $0.189$  & $0.179$  \\
\multicolumn{1}{l}{\cmark} &\cmark &\xmark &\xmark & $0.837$   & $0.582$  & $0.293$ \\
\multicolumn{1}{l}{\cmark} &\cmark &\cmark &\xmark & $\textbf{0.874}$   & $0.658$  & $0.327$ \\
\hline \multicolumn{1}{|l}{\cmark} &\cmark &\cmark &\cmark & $0.872$  & $\textbf{0.673}$  & \multicolumn{1}{c|} {$\textbf{0.348}$} \\ \hline
\end{tabular}
}
\caption{\label{table:ablation1} Ablations: impact of component combinations and weight-tying on F-score (higher is better). Our reported model is boxed.}
\end{table}

\paragraph{Evolution.} Figure~\ref{fig:evolution} shows an example of how the distance map $\bar{d}[n]$ evolves during refinement. Specifically, we visualize the result of slicing $b_k(x; \mathbf{g})$, to which we apply a non-linearity that amplifies less-prominent boundaries. We see that early iterations are exploratory and unstructured, and that later iterations agree.

\section{Ablations}\label{sec:ablations}

\Cref{table:ablation1} shows what happens when the initial per-patch Neighborhood MLP-mixer [NM] is used alone, versus combining it with one boundary attention block [BA-1] or two such blocks [BA-2]. Tying the attention weights across iterations provides a slight advantage at higher noise levels, with only a slight penalty in accuracy at lower noise. Our final model (boxed) provides the best overall performance, accepting slightly lower accuracy at low noise in exchange for better accuracy at higher noise levels.

\Cref{table:ablation2} compares many other variations of the boxed model from \cref{table:ablation1}, with asterisks indicating the model specifications used in the paper. \emph{Iterations} varies the number of attention-iterations within each block, and \emph{Neighborhood} varies the attention neighborhood size. \emph{MLP Input} varies the features concatenated to the hidden state prior to the pre-attention MLP and shows that replacing the gathered colors $\bar{\mathbf{f}}^t$ with a constant array or the input image values $\mathbf{f}$ performs worse. \emph{Windowing} shows that using fixed, square windowing functions $w_n(x)$ of any size performs equal (at low noise) or worse than inferring them adaptively. Somewhat intuitively, a small $9\times 9$ patch size performs well at low noise, but performance lags under noisy conditions where using a larger patch size increases the spatial extent of the communication across patches.

\begin{table}[t]
\scriptsize
\centering
\setlength{\tabcolsep}{6pt}
{
\centering
\begin{tabular}{l | l | c c c l}
\multicolumn{1}{l}{} & & \! Low Noise$\uparrow$   & \! Med-Noise$\uparrow$   & \! High Noise$\uparrow$   \\ \hline
\multirow{3}{*}{Iterations}
 &3 & $0.867$  & $0.647$  & $0.329$ \\
 &4* & $\textbf{0.872}$  & $\textbf{0.673}$  & {$\textbf{0.348}$} \\
 &5 & $0.871$  & $0.663$  & $0.337$ \\\hline 
\multirow{4}{*}{Neighborhood} 
 &$7\times 7$ & $0.869$   & $0.637$  & $0.313$  \\
 &$9\times 9$ & $0.871$   & $0.658$  & $0.325$  \\
 &$11 \times 11$* & $\textbf{0.872}$  & $\textbf{0.673}$  & $\textbf{0.348}$ \\
 &$13\times 13$ & $0.871$   & $0.655$  & $0.324$  \\ \hline
\multirow{3}{*}{MLP Input} & Constant & $0.870$   & $0.650$  & $0.317$  \\
 &Input features & $0.868$   & $0.657$  & $0.316$  \\
 &Avg.~features* & $\textbf{0.872}$  & $\textbf{0.673}$  & $\textbf{0.348}$ \\ \hline
\multirow{2}{*}{Windowing}
&Fixed & $\textbf{0.872}$  & $0.662$  & $0.328$ \\
 &Inferred* & $\textbf{0.872}$  & $\textbf{0.673}$  &$\textbf{0.348}$ \\
\end{tabular}
}
\caption{\label{table:ablation2} Ablations: variations of our reported model. We compare the number of iterations of boundary attention, the attention neighborhood size, the features concatenated to the hidden state as input to the MLP, and a fixed versus learned windowing function. Asterisks indicate our reported choices.}
\end{table}

\section{Conclusion}
We have introduced a differentiable model that uses boundary attention to explicitly reason about geometric primitives such as edges, corners, junctions, and regions of uniform appearance in images.
Our bottom-up, feedforward network can encode images of any resolution and aspect ratio into a field of geometric primitives that describe the local structure within every patch. This work represents a step to the goal of synergizing the benefits of low-level parametric modeling with the efficiency and representational power of deep learning.

\ifSubfilesClassLoaded{
\bibliographystyle{splncs04}
\bibliography{refs}
}{}

\end{document}

\bibliographystyle{splncs04}

\addtocontents{toc}{\protect\setcounter{tocdepth}{3}}

\newpage
\clearpage

\setcounter{section}{0}

\title{\large Supplemental Material for Boundary Attention: Learning curves, corners, junctions and grouping}
\titlerunning{Supplemental Material for Boundary Attention}
\author{}
\authorrunning{M. Polansky et al.}
\institute{}

\maketitle

\setcounter{tocdepth}{2}
\tableofcontents

\supsection{The space of $M$-junctions}

Here we provide the expressions for the support functions $s_{j}(x;\mathbf{g})$ and the unsigned distance function $d(x;\mathbf{g})$ from Section~3 of the main paper. We also describe the differences between our parameterization of junction space and the original one in the field of junctions~\cite{verbin2021field}, with the new parameterization's main advantages being the avoidance of singularities and the ability to define mechanisms for smooth interpolation. Our descriptions of these require introducing a few additional mathematical details. We provide these details for the general case of geometric primitives (junctions) $\mathbf{g}$ that have $M$ angular wedges $\boldsymbol{\omega}=(\omega_1,\ldots,\omega_M)$, for which the paper's use of $M=3$ is a special case.

To begin, consider a local region $\Omega(x)\subset\mathbb{R}^2$ and fix a positive integer value for the maximum number of angular wedges $M>0$ (the paper uses $M=3$). Our partitions are parameterized by $\mathbf{g}\in\mathbb{R}^2\times\mathbb{S}^1\times\triangle^{M-1}$, where $\mathbb{S}^1$ is the unit circle and $\triangle^{M-1}$ is the standard $(M-1)$-simplex (\ie, the set of $M$-vectors whose elements are nonnegative and sum to one). We use the notation $\mathbf{g}=(\boldsymbol{u}, \theta, \boldsymbol{\omega})$, where $\boldsymbol{u}=(u,v)\in \mathbb{R}^2$ is the \emph{vertex}, $\theta\in\mathbb{S}^1$ is the \emph{orientation}, and   $\boldsymbol{\omega}=(\omega_1,\omega_2,\ldots,\omega_M)$ are barycentric coordinates (defined up to scale) for the $M$ relative angles, ordered clockwise starting from $\theta$. As noted in the main paper, our convention is to express the vertex coordinates relative to the center of region $\Omega(x)$, and we note again that the vertex is free to move outside of this region. We also note that up to $M-1$ of the angles $\omega_j$ can be zero. When necessary, we use notation $\tilde{\boldsymbol{\omega}}=(\tilde\omega_1,\tilde\omega_2,\ldots,\tilde\omega_M)$ to represent angles that are normalized for summing to $2\pi$:
\begin{equation}\label{eq:normalized-wedges}
    \tilde{\boldsymbol{\omega}}=\frac{2\pi\boldsymbol{\omega}}{\sum_{j=1}^M\omega_j}.
\end{equation}

As an aside, we note that there are some equivalences in this parameterization. First, one can perform, for any $k\in \{1\ldots (M-1)\}$, a cyclic permutation of the angles $\boldsymbol{\omega}$ and adjust the orientation $\theta$ without changing the partition. That is, the partition does not change under the cyclic parameter map
\begin{align}
 \omega_j & \rightarrow \omega_{j+k(\textrm{mod} M)}\label{eq:cyclic-angle}\\
\theta & \rightarrow \theta-\sum_{j=M+1-k}^{M} \omega_{j}\label{eq:cyclic-orientation}
\end{align}
for any $k\in \{1\ldots (M-1)\}$. 
Also, an $M$-junction $\left(\boldsymbol{u},\theta,(\omega_1,\ldots,\omega_M)\right)$ provides the same partition as any $M'$-junction, $M'>M$, that has the same vertex and orientation along with angles $(\omega_1\ldots\omega_M,0\dots)$. This captures the fact that $M$-junction families are nested for increasing $M$.

\begin{supfigure}[H]
    \centering
    \includegraphics[width=\textwidth]{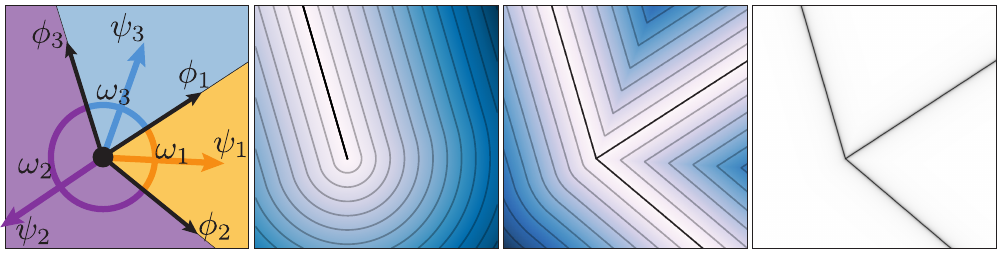}
    \caption{Anatomy of an $M$-junction $\mathbf{g}=(\mathbf{u},\theta,\boldsymbol{\omega})$ with $M=3$. \emph{Left}: Boundary directions $\phi_j$ and central directions $\psi_j$ are determined directly from relative angles $\boldsymbol{\omega}$ and orientation $\theta$ (which is equal to $\phi_1$). \emph{Middle panels}: Unsigned distance function for a boundary ray $d_3(x;\mathbf{g})$ and overall unsigned distance function $d(x;\mathbf{g})$, which is the minimum of the three per-ray ones. \emph{Right}: Associated boundary function $b_\eta(x;\mathbf{g})$ using $\eta=0.7$. \label{fig:junction-supp}}
\end{supfigure}

As shown in Figure~\ref{fig:junction-supp}, other geometric features of a junction can be directly derived from the orientation and angles. The \emph{central directions} $\boldsymbol{\psi}=(\psi_1,\ldots,\psi_M)$ are
\begin{equation}
    \psi_j = \theta 
    + \frac{\tilde{\omega}_j}{2} 
    + \sum_{k=1}^{j-1}\tilde{\omega}_k,\quad j\in\{1\ldots M\},
\end{equation}
and the \emph{boundary directions} $\boldsymbol{\phi}=(\phi_1,\ldots,\phi_M)$ are given by $\phi_1=\theta$ and
\begin{equation}
    \phi_j = \theta 
    + \sum_{k=1}^{j-1}\tilde{\omega}_k,\quad j\in\{2\ldots M\}.
\end{equation}

A key difference between our new parameterization of $M$-junctions and the original one~\cite{verbin2021field} is that the latter comprises $\left(\boldsymbol{u},\boldsymbol{\phi}\right)$ and requires enforcing constraints $0\le\phi_1\le\phi_2\le\cdots\le\phi_M\le2\pi$ (or somehow keeping track of the permutations of wedge indices that occur when these constraints are not enforced). The new $\left(\boldsymbol{u},\theta,\boldsymbol{\omega}\right)$-parameterization eliminates the need for such constraints.

As noted in the main paper's Section~3, we define the $j$th \emph{support} $s_{j}(x;\mathbf{g})$ as the binary-valued function that indicates whether each point $x\in\Omega$ is contained within wedge $j\in\{1\ldots,M\}$. Its expression derives from the inclusion condition that the dot product between the vector from the vertex to $x$ and the $j$th central vector  $\left(\cos\psi_j,\sin\psi_j \right)$ must be smaller than the cosine of half the angle $\tilde\omega_j$. Using Heaviside function $H(\cdot)$ we write
\begin{equation}\label{eq:wedge-indicator}
\begin{split}
s_j(x;\mathbf{g}) = 
H\Big((x-\mathbf{u})\cdot(\cos\psi_j,\sin\psi_j)
- \cos(\tilde{\omega}_j/2) ||x-\mathbf{u}|| \Big).
\end{split}
\end{equation}
As an aside, observe that this expression remains consistent for the case $M=1$, where there is a single wedge. In this case, $\tilde{\boldsymbol{\omega}}=\tilde{\omega}_1=2\pi$ by Equation~\ref{eq:normalized-wedges}, and the support reduces to $s_1(x)=1$ for all vertex and orientation values. 

The \emph{unsigned distance} $d(x;\mathbf{g})$ represents the Euclidean distance from point $x$ to the nearest point in the boundary set defined by $\mathbf{g}$. It is the minimum over $M$ sub-functions, with each sub-function being the unsigned distance from a boundary ray that extends from point $\mathbf{u}$ in direction $\phi_j$. The unsigned distance from the $j$th boundary ray is equal to the distance from its associated line for all points $x$ in its containing half-plane; and for other points it is equal to the radial distance from the vertex. That is,
\begin{equation}
    d_j(x;\mathbf{g}) = 
    \begin{cases}
    \left|(x-\mathbf{u})\cdot (-\sin\phi_j,\cos\phi_j)\right|,& \text{if } (x-\mathbf{u})\cdot (\cos\phi_j,\sin\phi_j) > 0\\
    \|(x-\mathbf{u})\|,&  \text{otherwise}.
    \end{cases}
\end{equation}
Then, the overall distance function is 
\begin{equation}
    d(x;\mathbf{g})=\min_{j\in{1\dots M}} d_j(x;\mathbf{g}).
\end{equation}

Finally, analogous to Equation~7 in the main paper, we define a junction's boundary function $b_\eta(x;\mathbf{g})$ as the result of applying a univariate nonlinearity to the unsigned distance:
\begin{equation}
    b_\eta(x;\mathbf{g})=\left(1 + (d(x;\mathbf{g})/\eta)^{2}\right)^{-1}.
\end{equation}
Figure~\ref{fig:junction-supp} shows an example of a junction's distance function and its associated boundary function with $\eta=0.7$.

\supsubsection{Interpolation}
Another advantage of the present parameterization compared to that of the original~\cite{verbin2021field} is that it is a simply-connected topological space and so allows for defining mechanisms for smoothly interpolating between any two junctions $\mathbf{g}=\left(\boldsymbol{u},\theta,\boldsymbol{\omega}\right)$ and $\mathbf{g}'=\left(\boldsymbol{u}',\theta',\boldsymbol{\omega}'\right)$. In our implementation we define interpolation variable  $t\in[0,1]$ and compute interpolated junctions $\mathbf{g}(t)=\{\boldsymbol{u}(t), \theta(t), \boldsymbol{\omega}(t)\}$ using
a simple combination of linear and spherical linear interpolation: 
\begin{align}
    \boldsymbol{u}(t) & = (1-t)\boldsymbol{u} + t\boldsymbol{u}' \\
    \tilde{\boldsymbol{\omega}}(t) & = (1-t)\tilde{\boldsymbol{\omega}} + t\tilde{\boldsymbol{\omega}}',
\end{align}
and
\begin{equation}
    \theta(t) = \theta + t\Delta\theta,
\end{equation}
with
\begin{equation*}
    \Delta\theta = 
    \begin{cases}
    \theta' - \theta - 2\pi, & \text{if } \theta' - \theta > \pi  \\
    \theta' - \theta + 2\pi, & \text{if } \theta' - \theta < -\pi \\
    \theta' - \theta, & \text{otherwise},
    \end{cases}
\end{equation*}
assuming $\theta,\theta'\in[0,2\pi)$. The bottom row of Figure~7 in the main paper visualizes a set of samples from smooth trajectories in junction space using this mechanism.

\supsection{Training Data}

\begin{supfigure}[H]
\center
\includegraphics[width=.6\columnwidth]{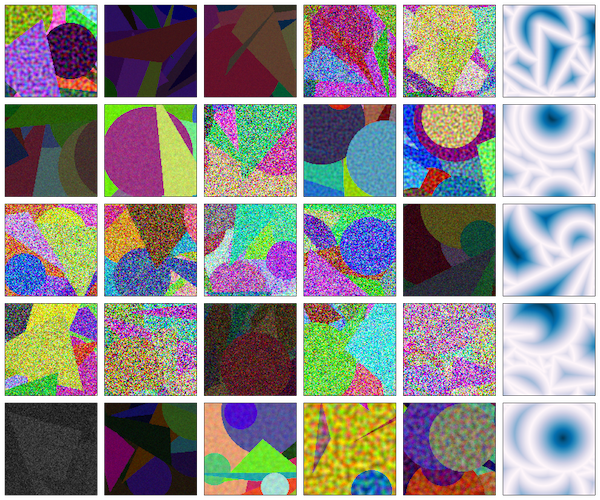}
\caption{\label{fig:trainingdata} \emph{Columns 1 to 5:} Examples of the synthetic data used to train our model using supervision with ground-truth boundaries. \emph{Column 6:} Rendered distance maps corresponding to column 5. The training data contains random circles and triangles that each have a random RGB color, and the images are corrupted by various types and amounts of noise. Each noiseless image has an unrasterized, vector-graphics representation of its shapes and colors, which specify the clean image and exact boundary-distance map with unlimited resolution. 
}
\end{supfigure}

We find that we can train our model to a useful state using purely synthetic data, examples of which are depicted in Figure~\ref{fig:trainingdata}. In fact, we find it sufficient to use very simple synthetic data that consists of only two basic shapes---circles and triangles---because these can already produce a diverse set of local edges, thin bars, curves, corners, and junctions, in addition to uniform regions. We generate an image by randomly sampling a set of circles and triangles with geometric parameters expressed in continuous, normalized image coordinates $[0,1]\times[0,1]$. We then choose a random depth ordering of the shapes, and we choose a random RGB color for each shape. Importantly, the shape and color elements are specified using a vector-graphics representation, and the shape elements are simple enough to provide an exact, symbolic expression for each image's true boundary-distance map, without approximation or rasterization. They also allow calculating the precise locations, up to machine precision, for all of the visible corners and junctions in each image.

At training time, an input image is rasterized and then corrupted by a random amount and type of noise, including some types of noise that are spatially-correlated. This forces our model to only use color as its local cues for boundaries and grouping; and it forces it to rely heavily on the topological and geometric structure of curves, corners and junctions, as well as their contrast polarities. The highly-varying types and amounts of noise also encourages the model to use large window functions $w(x;\mathbf{g})$ when possible, since that reduces noise in the gather operation and reduces variance $\nu_f[n]$.

Our dataset, which we call Kaleidoshapes, is available publicly, along with the code for generation, training and evaluation.

\boldstart{Shapes and colors.} For our experiments, we rasterized each image and its true distance map at a resolution of $240 \times 320$ images, with each one containing between 15 and 20 shapes. We used a $40\!\!:\!\!60$ ratio of circles to triangles. In terms of normalized coordinates, circles had radii in the range $[0.05,0.2]$ and triangles had bases in the range $[0.02,0.5]$ and heights in the range $[0.05,0.3]$. This allows triangles to be quite thin, so that some of the local regions $\Omega(x)$ contain thin bar-like structures. Additionally, we included a minimum visibility threshold, filtering out any shapes whose visible number of rasterized pixels is below a threshold. Colors were selected by uniformly sampling all valid RGB colors. During training, batches consisted of random $125 \times 125$ crops.

\boldstart{Noise.} For noise types, we used combinations of additive zero-mean Gaussian noise; spatially average-pooled Gaussian noise; Perlin noise~\cite{perlin1985image}, and simulated photographic sensor noise using the simplified model from~\cite{wei2020physicsbased}. The total noise added to each image was sampled uniformly to be between 30\% and 80\% of the maximum pixel magnitude, and then noise-types were randomly combined with associated levels so that they produced the total noise level. Since zero-mean noise can at times result in values below 0 or above the maximum magnitude threshold, we truncate any pixels outside of that range.

\supsection{Model Details}

Our model is designed to be purely local and bottom up, with all of its compositional elements operating on spatial neighborhoods in a manner that is invariant to discrete spatial shifts of an image. Its design also prioritizes having a small number of learnable parameters. Here we provide the details of the two blue blocks in the main paper's Figure~3: Neighborhood MLP-Mixer and Neighborhood Cross-attention. Our model was implemented with JAX and its code and trained weights are available publicly. 

\supsubsection{Neighborhood MLP-Mixer}

Our neighborhood MLP-mixer is a shift invariant, patch-based network inspired by MLP-mixer~\cite{tolstikhin2021mlpmixer}. It replaces the image-wide operations of~\cite{tolstikhin2021mlpmixer} with patch-wise ones. Given an input image, we first linearly project its pixels from $\mathbb{R}^3$ to dimension $\mathbb{R}^{D_\gamma}$ (we use $D_\gamma=64$), which is followed by two neighborhood mixing blocks. Each neighborhood mixing block contains a spatial patch mixer followed by a channel mixer. The spatial patch mixer is implemented as two $3 \times 3$ spatial convolutions with weights tied across channels. It thereby combines spatial patches of features with all channels (and patches) sharing the same weights. Following~\cite{tolstikhin2021mlpmixer}, we use GELU~\cite{DBLP:journals/corr/HendrycksG16} activations. The channel mixer is a per-pixel MLP with spatially-tied weights. To handle border effects in our neighborhood MLP-mixer, we apply zero-padding after the initial projection from $\mathbb{R}^{3}$ to $\mathbb{R}^{64}$, and then we crop to the input image size after the second neighborhood mixing block to remove features that correspond to patches without full coverage, \ie, patches that contain pixels outside of the original image.

\supsubsection{Neighborhood Cross-attention}

The neighborhood cross-attention block similarly enforces shift-invariance and weight sharing across spatial neighborhoods. Inside this block are two transformer layers whose cross-attention components are replaced with neighborhood cross-attention components that are restricted to a spatial neighborhood of pixels. We use $11\times 11$ neighborhoods in our implementation, which our ablations showed produced the best performance. In each neighborhood containing a query token, we add a learned positional encoding to the key/value tokens which is relative to the neighborhood's center and is the same for all neighborhoods. Then the query is updated using standard cross-attention with its neighborhood of key/values. We use 4 cross-attention heads.  Like the standard transformer, each neighborhood cross attention component is followed by an MLP, dropout layer, and additive residual. To handle border effects, we zero-pad the key and value tokens so that every query attends to an $11 \times 11$ neighborhood, and then zero-out any attention weights involving zero-padded tokens.

\supsubsection{Training Details}

During the first two stages of training where we train our model on junction images and simplified circle triangle images, we omit the global losses of Equations~$9$ and $10$. This primes the network to learn meaningful hidden states $\boldsymbol{\gamma}[n]$ and prevents the ``collapsing'' of junctions, where the boundary-consistency loss (\ie the sum over pixels of variance of distance $\nu_d[n]$) dominates and the network learns to predict all-boundaryless patches that are globally consistent but inaccurate. Because of data imbalance---only a small fraction of regions $\Omega_n(x)$ contain corners or junctions---we add an additional spatial importance mask to prioritize the regions that contain a corner (\ie, a visible triangle vertex) or a junction (\ie, an intersection between a circle and a triangle's edge). Our data generation process produces a list of all non-occluded vertices and intersections in each image, and we use these values to create a spatial importance mask with gaussians centered at each of these points. In practice, we use gaussians with a standard deviation of 7 pixels. This mask is added to the loss constant $C$.

The final stage of training adds a second boundary attention block with weights that are initialized using a copy of the pretrained weights of the first boundary attention block. We use $100k$ crops of size $125 \times 125$ from our Kaleidoshape images (10\% withheld for testing) and the full set of losses; and we optimize all of the model's parameters, including those of the neighborhood MLP-mixer and the first boundary attention block.  Like in pretraining, we add a spatial importance that prioritizes region containing a corner (\ie, a visible triangle vertex) or a junction (\ie, a visible intersection between the boundaries of any two shapes).

\supsection{Model Behavior}

\supsubsection{Qualitative Results for Natural Images}

\begin{supfigure}[t]
\center
\includegraphics[width=\columnwidth]{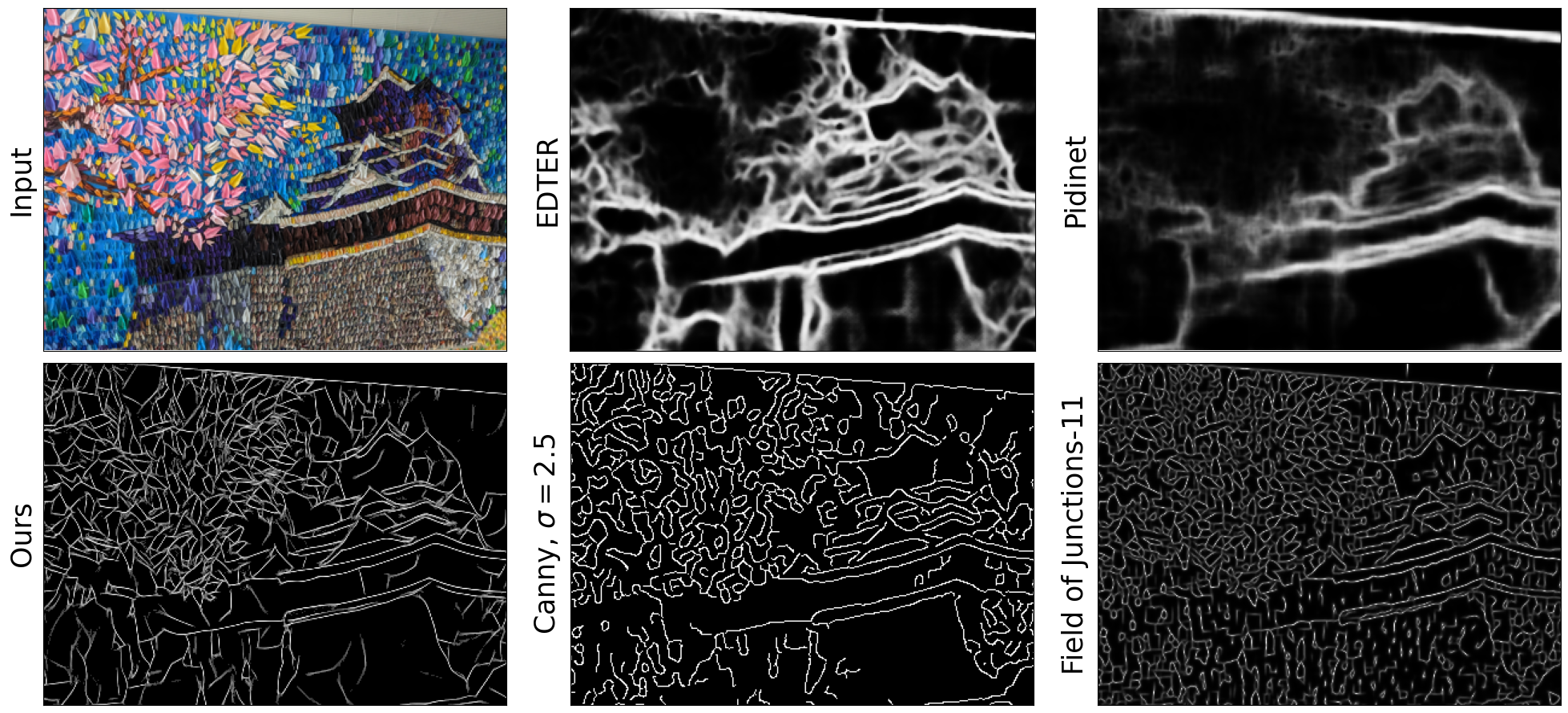}
\caption{\label{fig:machecomparison} Qualitative behavior of our model's output boundaries $\bar{b}_\eta[n]$ on noiseless natural images, compared to those of end-to-end models EDTER~\cite{pu2022edter} and Pidinet~\cite{su2021pixel} that are trained to match human annotations; and compared to two bottom-up methods that, like our model, are not trained to match human annotations: Canny~\cite{canny1986computational}, and the field of junctions (FoJ)~\cite{verbin2021field} with patch size $11$.
}
\end{supfigure}

In Figures~\ref{fig:japanimages} and~\ref{fig:machecomparison}, we show how the model behaves on noiseless natural images that contain texture and recognizable objects. In particular, Figure~\ref{fig:machecomparison} emphasizes how the boundary maps produced by our model qualitatively differ from other methods. Here, in addition to showing the results those of classical bottom-up edge-detectors, we include results learned, end-to-end models that have been trained to match human annotations as another point of reference. It is important to remember that these methods are trained to identify boundary structures that are defined by semantic variations, whereas our method and other low-level methods divide regions based on variations in color.

Figure ~\ref{fig:machecomparison} compares our output to that from Canny~\cite{canny1986computational}, the field of junctions (FoJ)~\cite{verbin2021field} with a patch size of $11$, Pidinet~\cite{su2021pixel}, and EDTER~\cite{pu2022edter}, the latter two being networks trained on human annotated data. (Note that inputs for all models besides EDTER~\cite{pu2022edter} were $300 \times 400$. Input to EDTER was down-sampled to $225 \times 300$ due to its input size constraint.)

We find that our model produces finer structures than the end-to-end learned models~\cite{dollar2014fast,pu2022edter} because it is trained to only use local spatial averages of color as its cue for boundaries and grouping. It does not include mechanisms for grouping based on local texture statistics, nor based on non-local shape and appearance patterns that have semantic meaning to humans. Compared to the bottom-up methods of Canny~\cite{canny1986computational} and the field of junctions~\cite{verbin2021field}, our model has the advantage of automatically adapting the sizes of its output structures across the image plane, through its prediction of windowing field $\mathbf{p}[k]$. In contrast, the the field of junctions and Canny both operate at a single pre-determined choice of local size, so they tend to oversegment some places while undersegmenting others. 

\begin{supfigure}[t]
\center
\begin{subfigure}{\textwidth}
\includegraphics[width=\columnwidth]{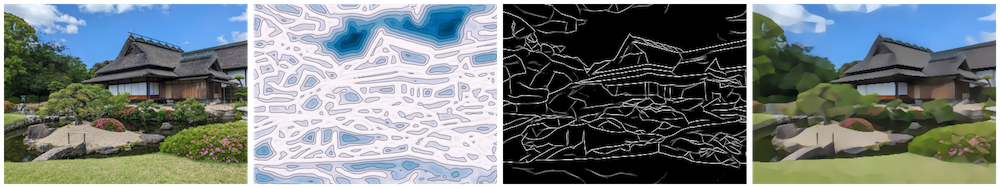}
\end{subfigure}
\begin{subfigure}{\textwidth}
\includegraphics[width=\columnwidth]{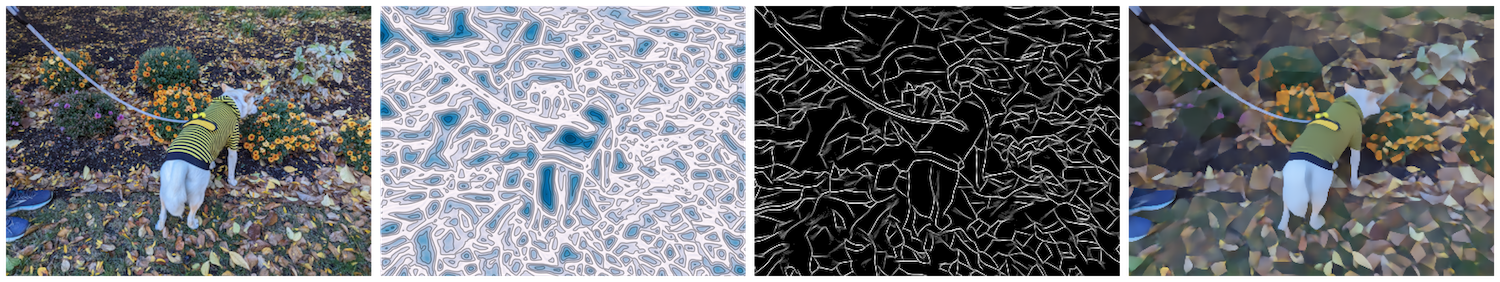}
\end{subfigure}
\begin{subfigure}{\textwidth}
\includegraphics[width=\columnwidth]{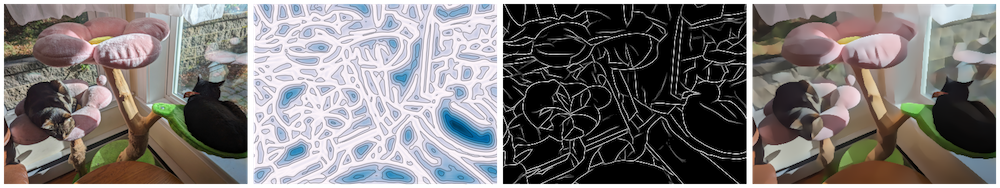}
\end{subfigure}
\caption{\label{fig:japanimages} Qualitative behavior of our model on noiseless natural images. \emph{From left to right:} Input image $\mathbf{f}[n]$, output distance map $\bar{d}[n]$, output boundary map $\bar{b}[n]$, and output boundary-smoothed features $\bar{\mathbf{f}}[n]$.}
\end{supfigure}

\supsection{Additional Examples for Low-light Images}

Figure~\ref{fig:reallowlight} shows examples of applying our model to indoor images taken by an iPhone XS in low light conditions. 

Figure~\ref{fig:pinwheelcomparison} provides additional comparisons for a sample of varying-noise images from the ELD dataset~\cite{wei2020physicsbased}. We include results from other low level methods and as a point of comparison, the outputs of several methods trained to parse semantic boundaries.

When detecting boundaries at low signal-to-noise ratios, it is difficult to accurately discern finer structures as the noise level increases. Some algorithms, such as the field of junctions (FoJ)~\cite{verbin2021field}, have tunable parameters such as patch-size that provide control over the level of detection. A small patchsize allows recovering fine structures in lower noise situations, but it causes many false positive boundaries at high noise levels. Conversely, a large patchsize provides more resilience to noise but has not ability to recover fine structure at all. Our model reduces the severity of this trade-off by automatically adapting its local windowing functions in ways that have learned to account for both the amount of noise and the local geometry of the underlying boundaries.

In Figure~\ref{fig:pinwheelcomparison} we see that our model is able to capture the double-contour shape of the curved, thin black bars, and that it continues to resolve them as the noise level increases, more than the other low-level methods. We also note that only the low-level models resolve this level of detail in the first place: The models trained on human annotations---UAED~\cite{zhou2023treasure}, EDTER, HED, Pidinet, and Structured Forests---miss the double contour entirely, estimating instead a single thick curve. We emphasize again that a user can adjust the behavior of Canny and the field of junctions by tuning their local size parameters, either the filter size for Canny or the patchsize for the field of junctions. Increasing the local size improves their resilience to noise but reduces their spatial precision. Neither system provides the ability to estimate fine grained details \emph{and} withstand noise, like our model does.

Figure~\ref{fig:extremenoise} contains additional examples of images cropped from the ELD dataset. Here we include examples with even higher levels of noise to show the complete degradation of our algorithm and others. 

\newpage

\begin{supfigure}[ht]
\begin{subfigure}{\textwidth}
\includegraphics[width=\columnwidth]{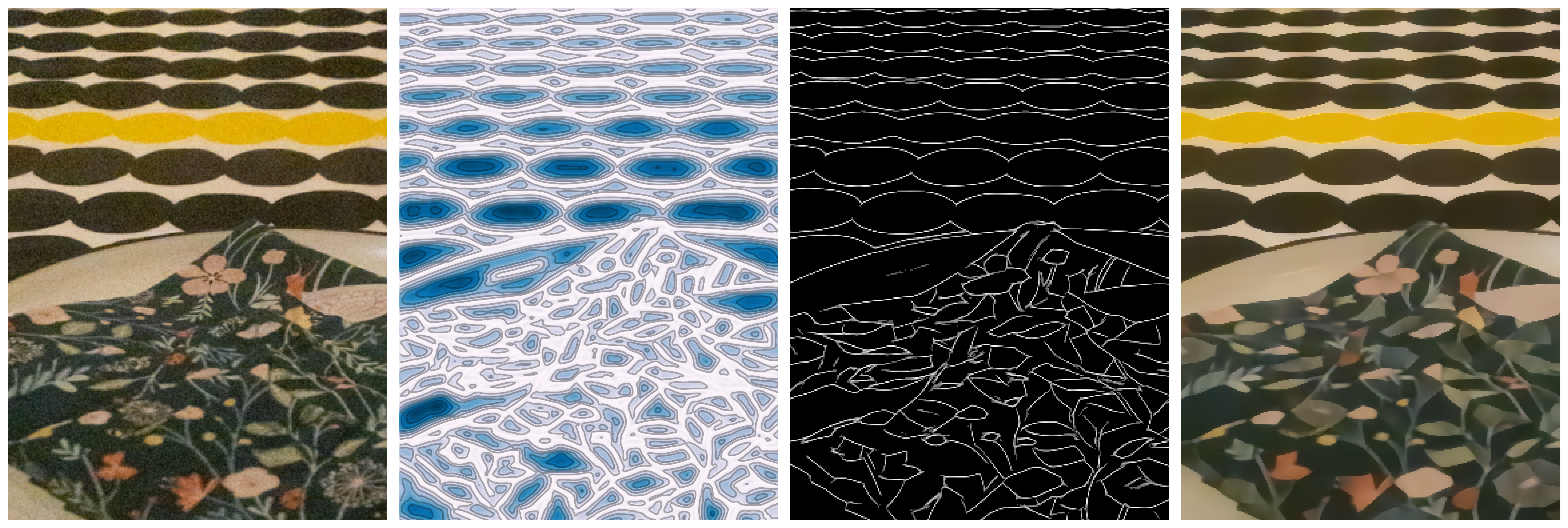}
\end{subfigure}
\begin{subfigure}{\textwidth}
\includegraphics[width=\columnwidth]{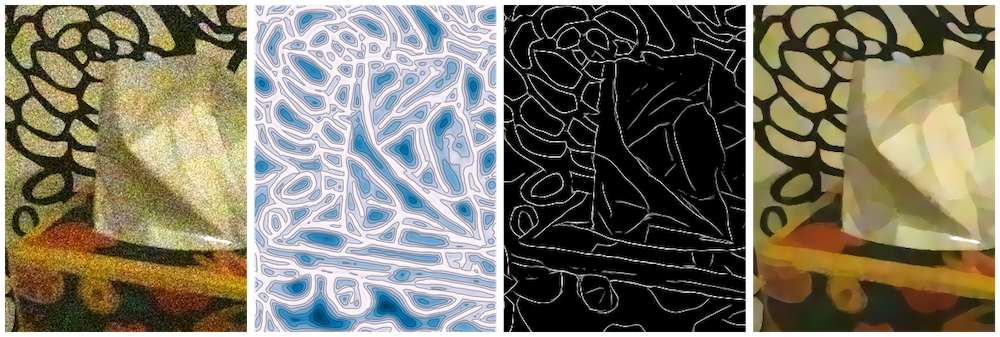}
\end{subfigure}
\begin{subfigure}{\textwidth}
\includegraphics[width=\columnwidth]{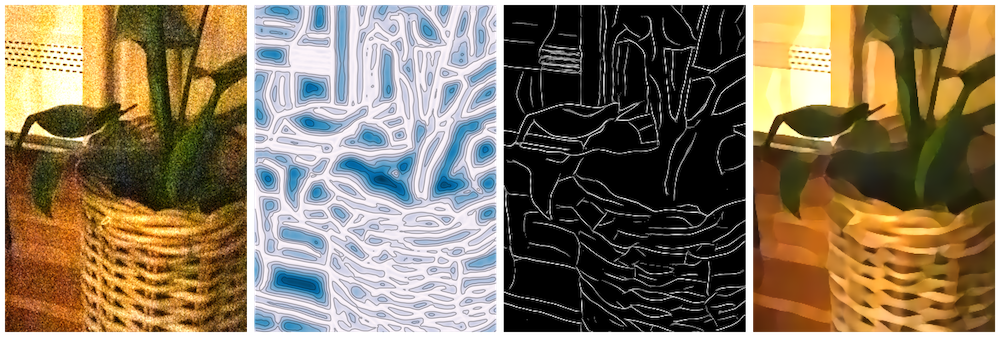}
\end{subfigure}
\caption{\label{fig:reallowlight} Visualization of our model's output for low-light images captured by an iPhone XS. \emph{From left to right:} Input image $\mathbf{f}[n]$, output distance map $\bar{d}[n]$, output boundary map $\bar{b}_\eta[n]$ with $\eta=0.7$, and output boundary-smoothed features $\bar{\mathbf{f}}[n]$.}
\end{supfigure}

\newpage
\begin{supfigure}[H]
\center
\includegraphics[width=\columnwidth]{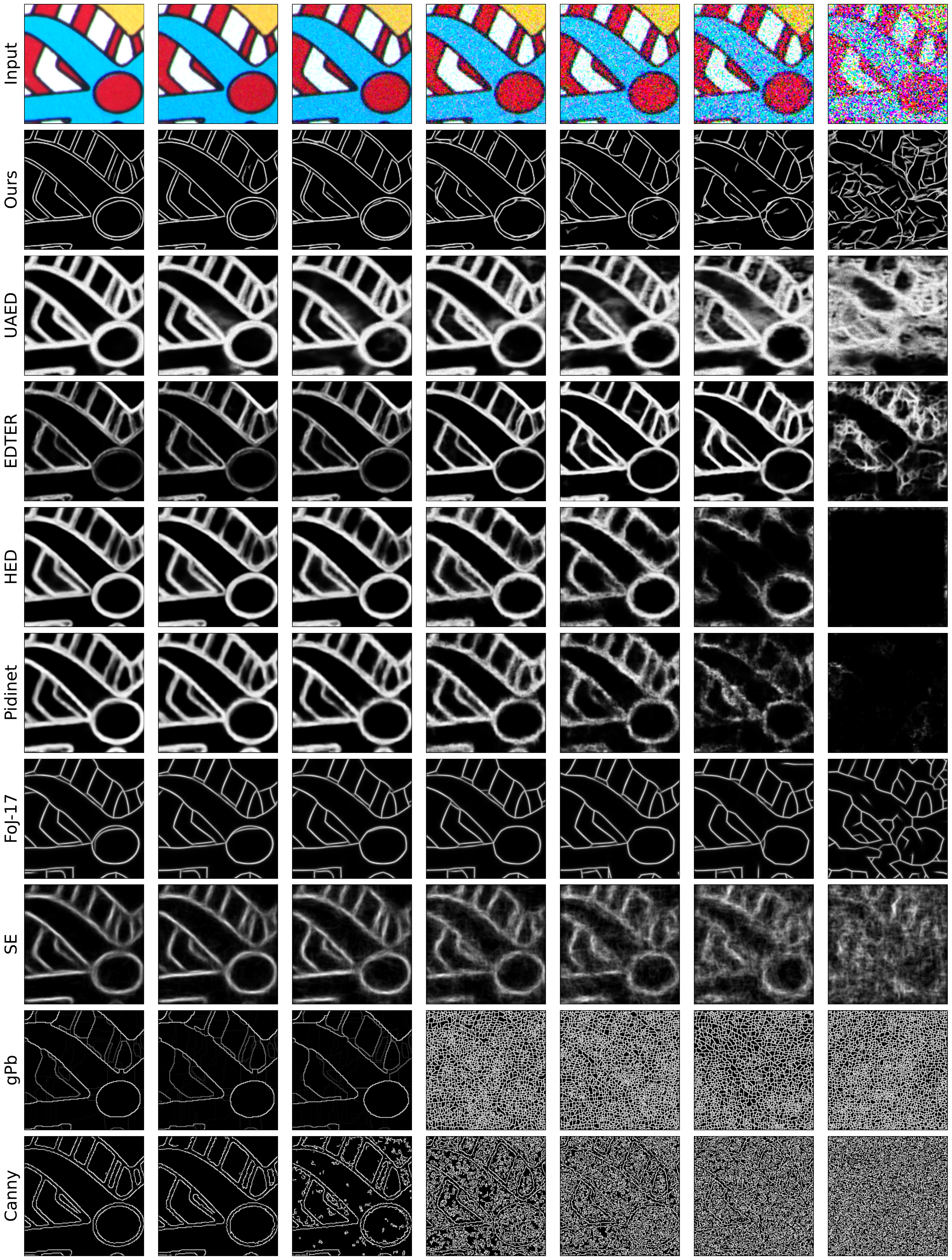}
\caption{\label{fig:pinwheelcomparison} Qualitative comparison between our model's output boundaries $\bar{b}_\eta[n]$ and those of other methods, for a crop from the ELD dataset under increasing amounts of photographic noise. We compare to end-to-end models that are trained to match human annotations (UAED~\cite{zhou2023treasure}, EDTER~\cite{pu2022edter}, HED~\cite{xie2015holistically}, Pidinet~\cite{su2021pixel}, and Structured Edges (SE)~\cite{dollar2014fast}) in addition to low-level models that are not (Canny~\cite{canny1986computational}, gPb~\cite{4587420}, and the field of junctions (FoJ-17)~\cite{verbin2021field}) with patch size 17. 
}
\end{supfigure}

\newpage

\begin{supfigure}[H]
\center
\begin{subfigure}{.48\textwidth}
\includegraphics[width=\columnwidth]{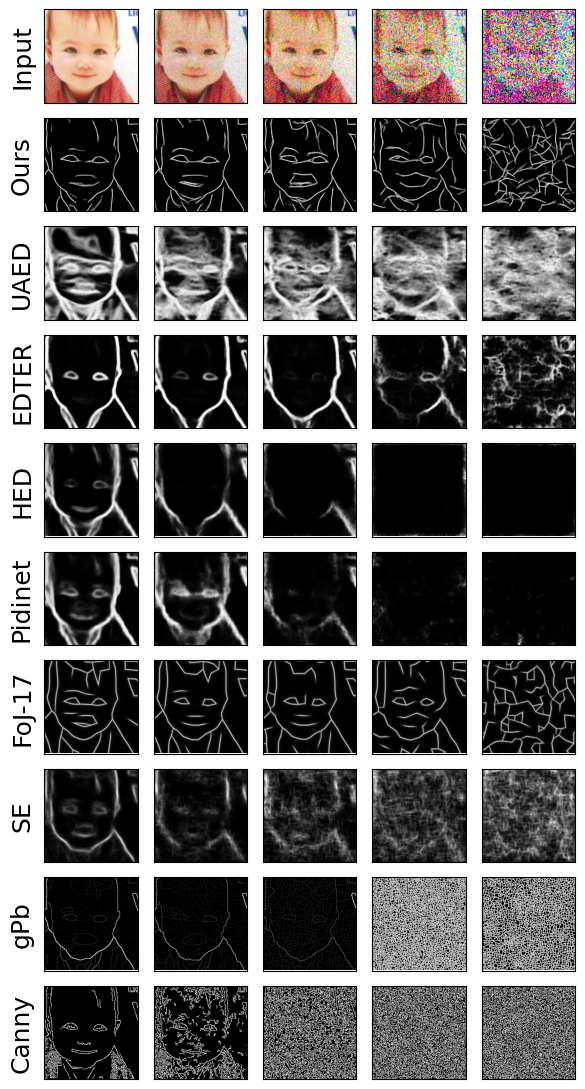}
\end{subfigure}
\begin{subfigure}{.48\textwidth}
\includegraphics[width=\columnwidth]{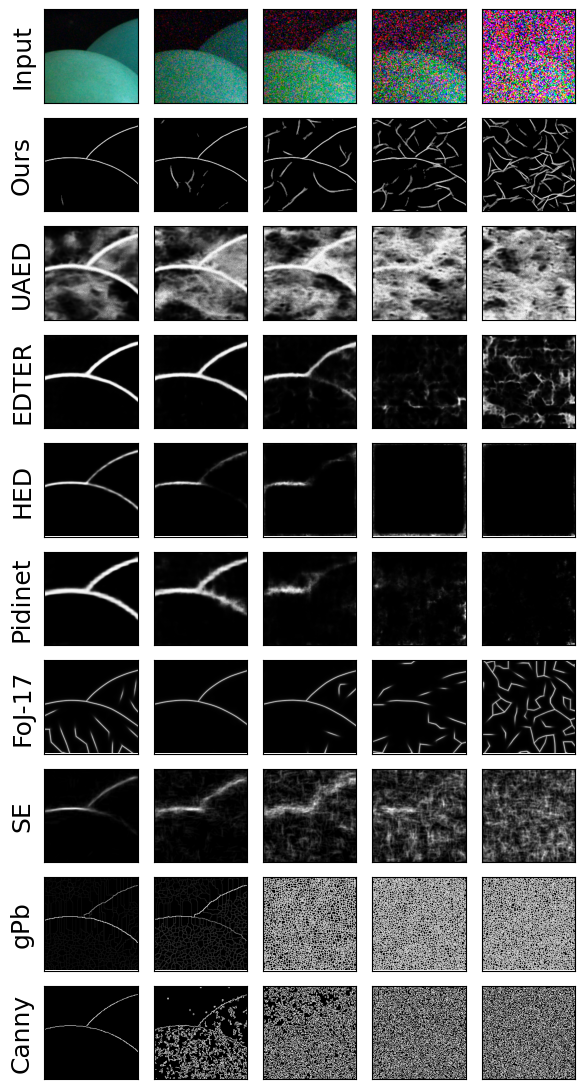}
\end{subfigure}
\caption{\label{fig:extremenoise}}
\end{supfigure}

\newpage

\addtocounter{suppfigure}{-1}
\begin{supfigure}[H]\ContinuedFloat
\center
\begin{subfigure}{.48\textwidth}
\includegraphics[width=\columnwidth]{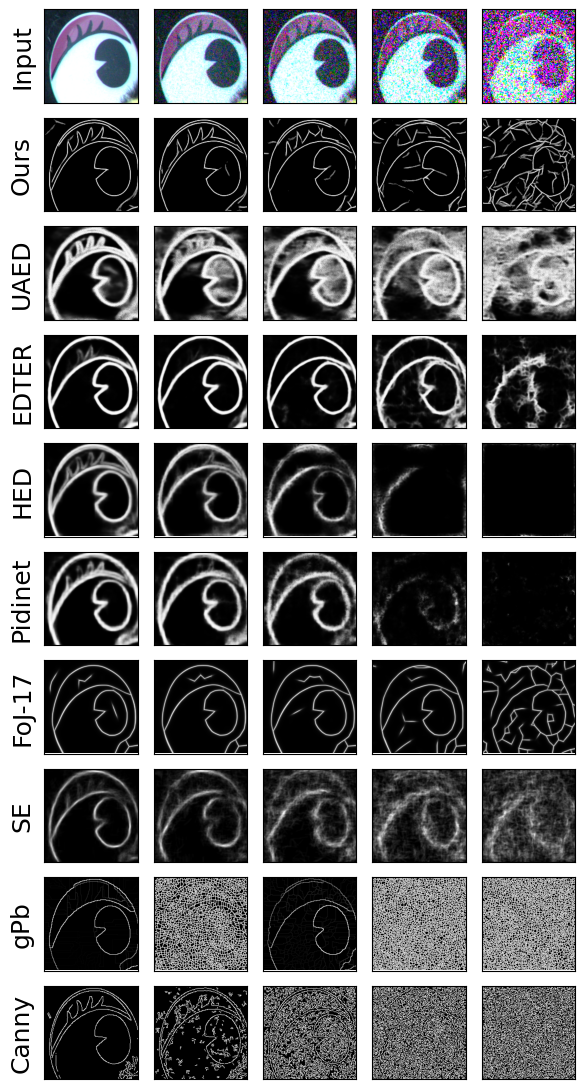}
\end{subfigure}
\begin{subfigure}{.48\textwidth}
\includegraphics[width=\columnwidth]{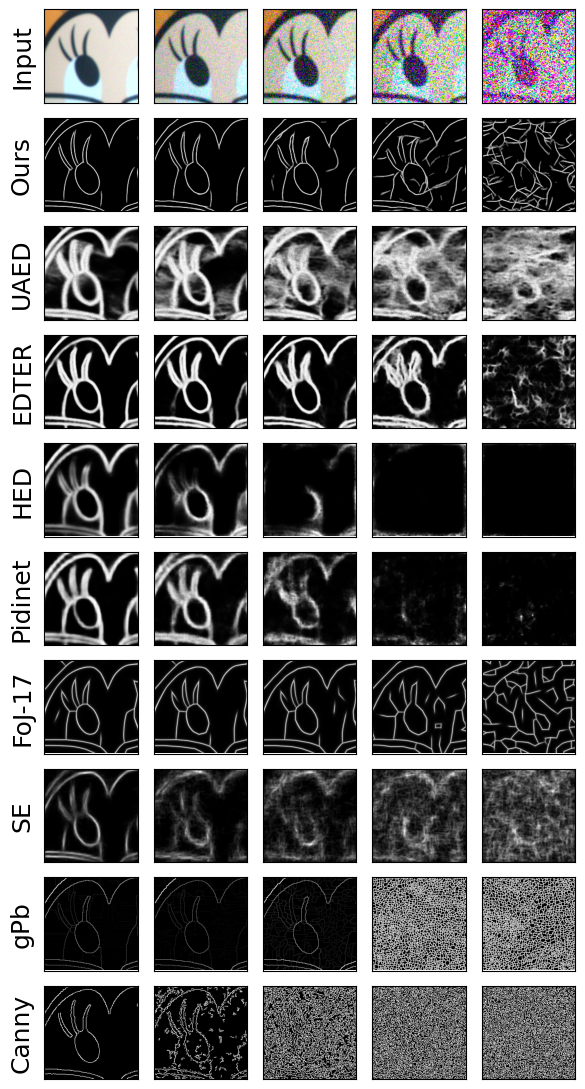}
\end{subfigure}
\caption{\label{fig:extremenoise} \emph{(cont.)}}
\end{supfigure}

\addtocounter{suppfigure}{-1}
\begin{supfigure}[H]\ContinuedFloat
\center
\begin{subfigure}{.48\textwidth}
\includegraphics[width=\columnwidth]{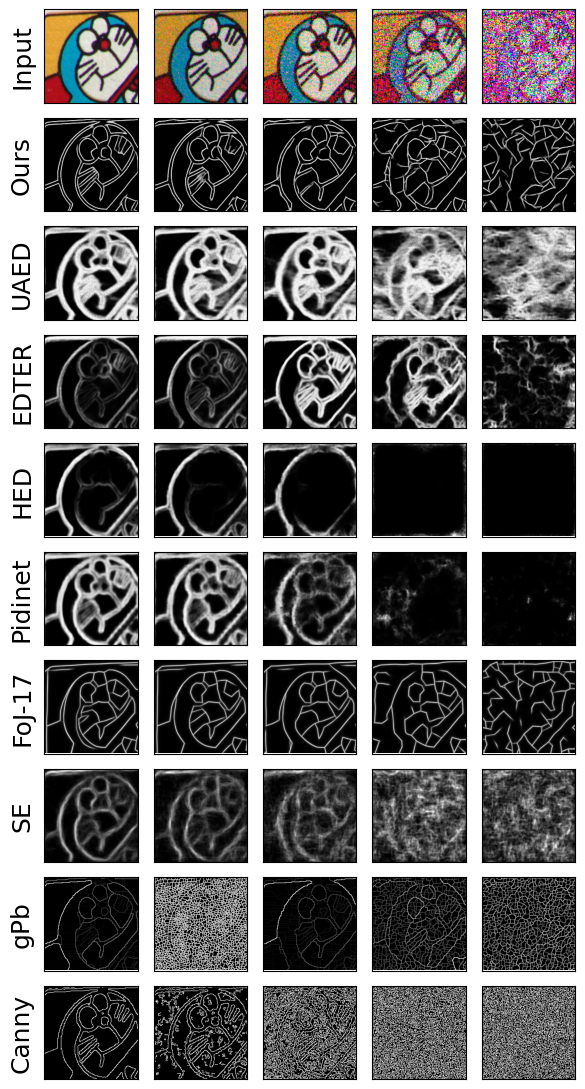}
\end{subfigure}
\begin{subfigure}{.48\textwidth}
\includegraphics[width=\columnwidth]{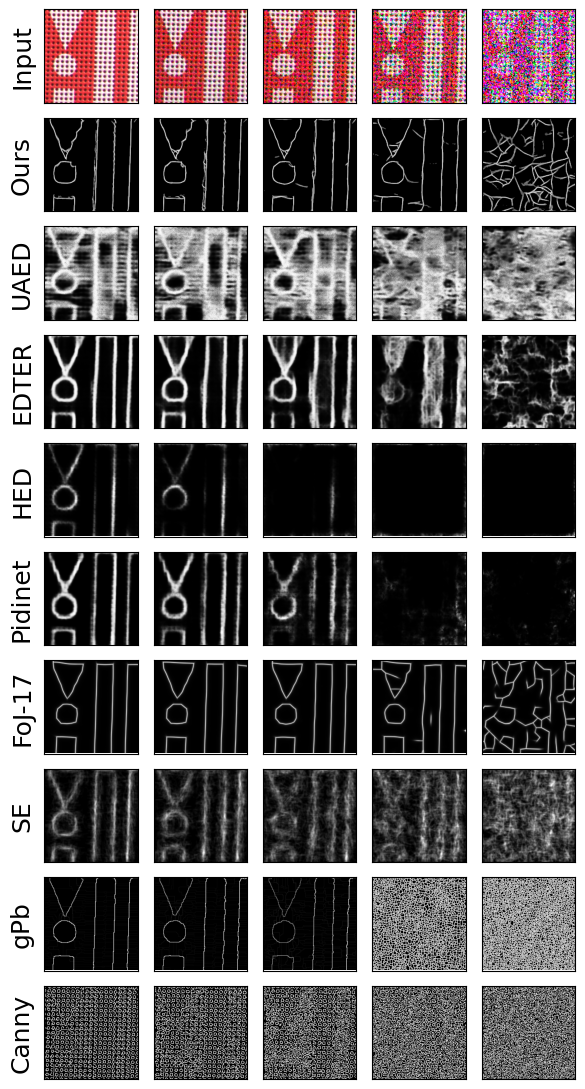}
\end{subfigure}
\caption{\label{fig:extremenoise} \emph{(cont.)}}
\end{supfigure}

\addtocounter{suppfigure}{-1}
\begin{supfigure}[H]\ContinuedFloat
\center
\begin{subfigure}{.48\textwidth}
\includegraphics[width=\columnwidth]{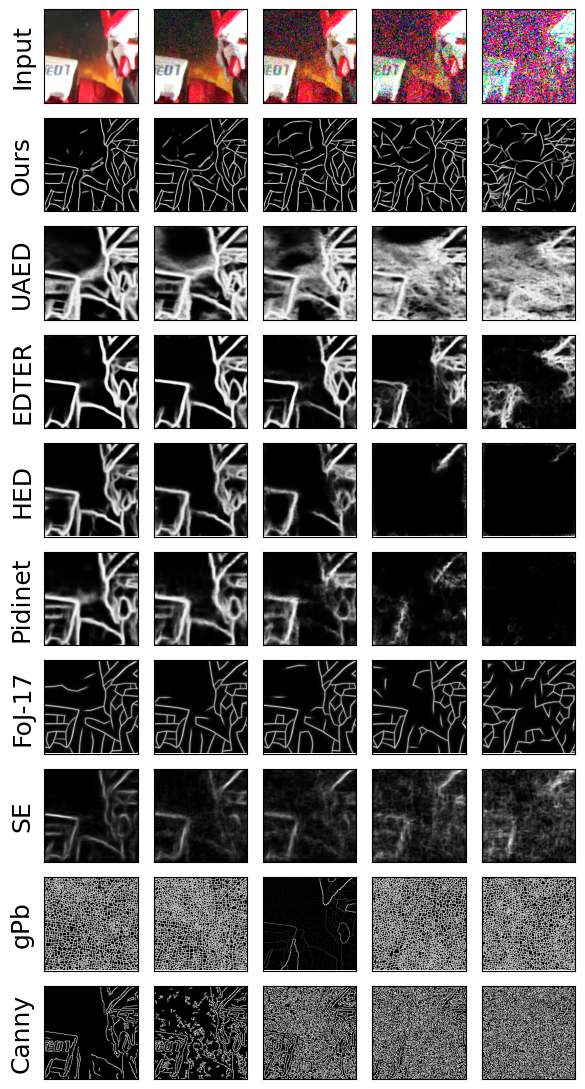}
\end{subfigure}
\begin{subfigure}{.48\textwidth}
\includegraphics[width=\columnwidth]{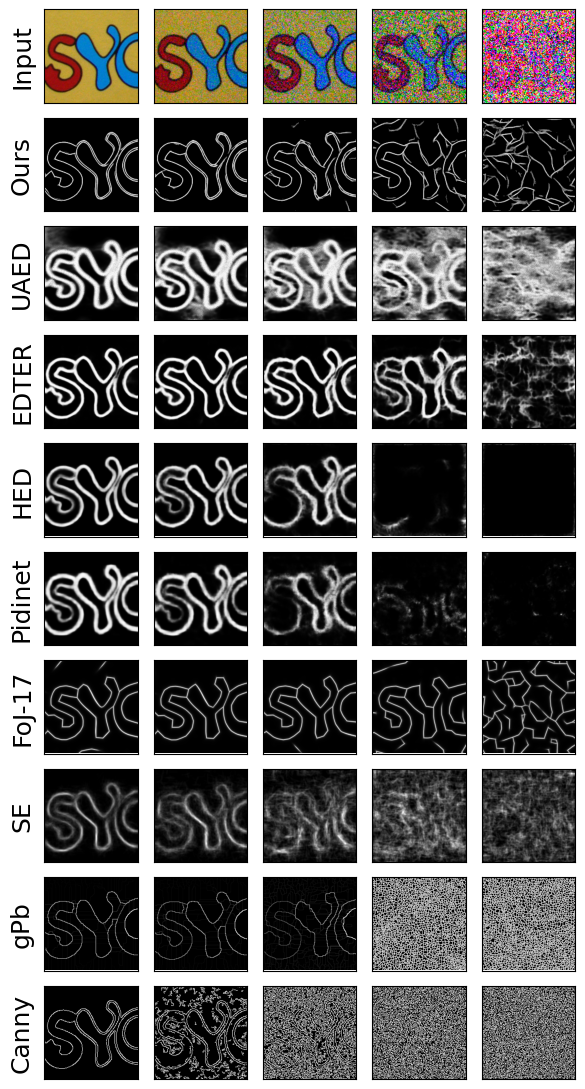}
\end{subfigure}
\caption{\label{fig:extremenoise} \emph{(cont.)} Additional qualitative comparisons between our model's output boundaries $\bar{b}_\eta[n]$ and those of other methods, using crops from the ELD dataset under increasing amounts of photographic noise, including very high levels of noise.}
\end{supfigure}

\supsection{Additional Uses of Our Model}

Here we demonstrate to uses of our model that follow directly from its output: hole-filling in RGBD images and non-photorealistic stylization.

\supsubsection{Color-based Depth Completion}

Figure~\ref{fig:depthcompletion} shows an example of using our model for simple hole-filling in the depth channels of RGBD images from the Middlebury Stereo Datasets~\cite{middlebury1,middlebury2}. We run our model on the RGB channels, and then for each pixel $n$ that has a missing depth value, we use our model's output local attention kernels $a_n(x)$ to fill in that pixel's value using an attention-weighted average of the observed depth values around it. This simple algorithm can be applied whenever the hole sizes are smaller than the maximum diameter of our attention maps, which is $34\times 34$ in our current implementation).

\newpage

\begin{supfigure}[H]
\begin{subfigure}{\textwidth}
\includegraphics[width=\columnwidth]{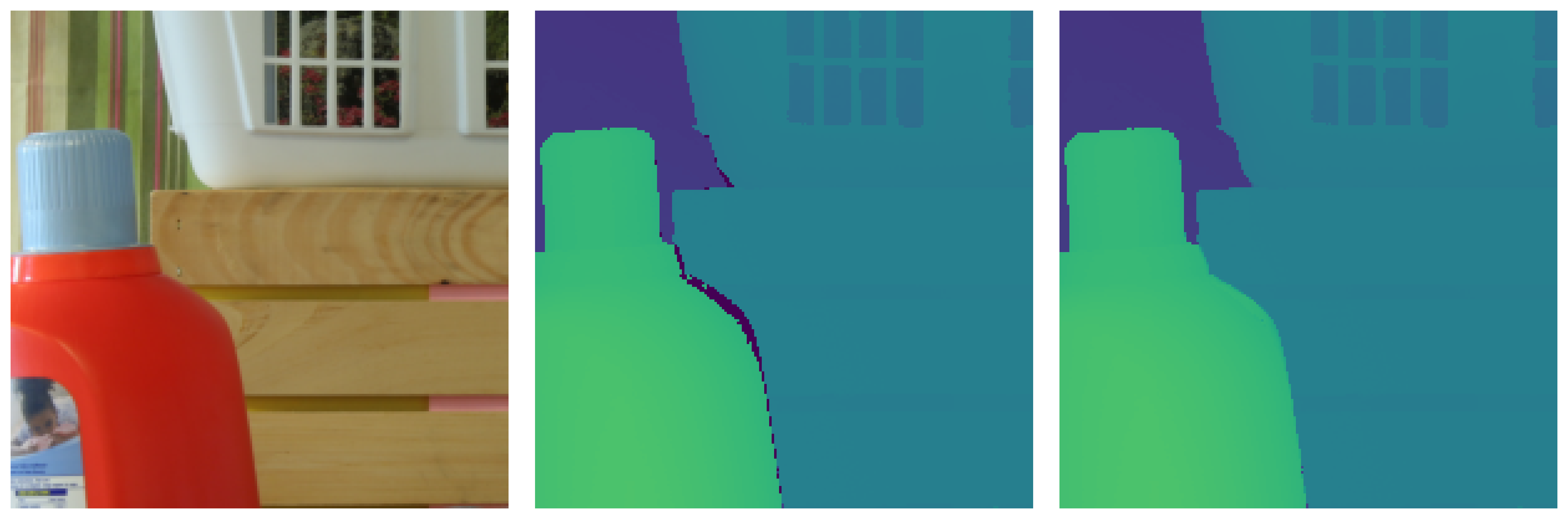}
\end{subfigure}
\begin{subfigure}{\textwidth}
\includegraphics[width=\columnwidth]{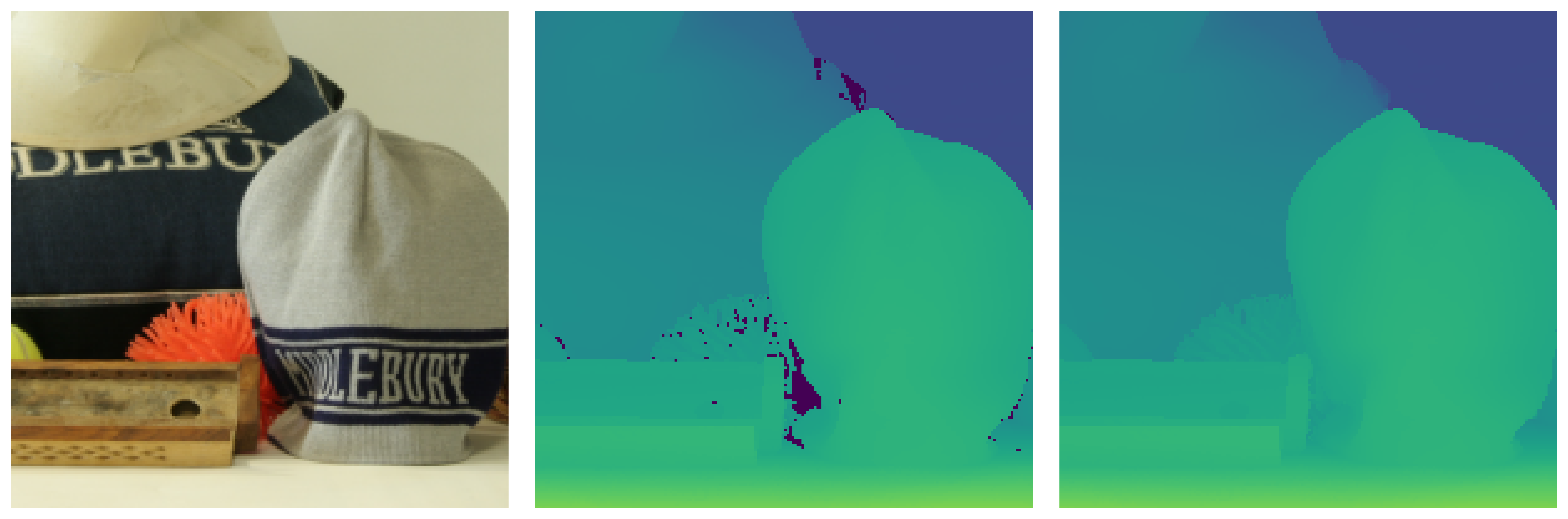}
\end{subfigure}
\begin{subfigure}{\textwidth}
\includegraphics[width=\columnwidth]{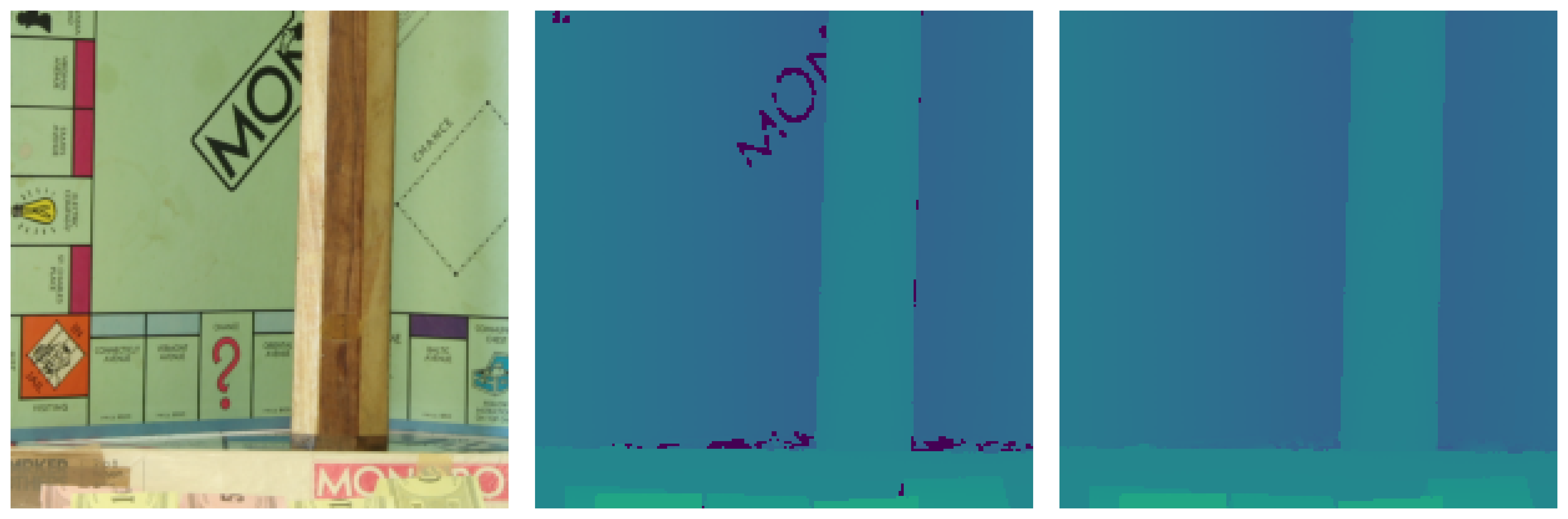}
\end{subfigure}
\begin{subfigure}{\textwidth}
\includegraphics[width=\columnwidth]{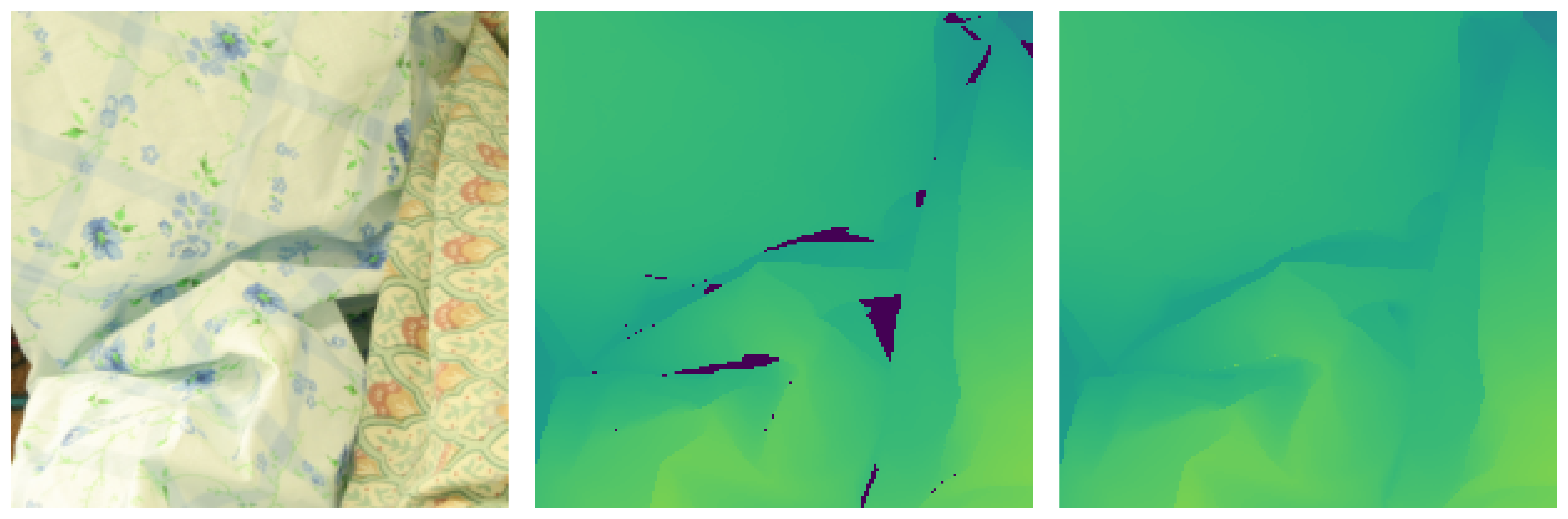}
\end{subfigure}
\caption{\label{fig:depthcompletion} Using our model for depth completion in RGBD images. \emph{Left:} Input RGB channels. \emph{Middle:} Input depth channel, with dark blue indicating missing values. \emph{Right:} Completed depth using our model's output attention kernels.}
\end{supfigure}

\newpage

\supsubsection{Application: Photo Stylization}

Figure~\ref{fig:stylization} shows examples of using our model's output for image stylization, by superimposing an inverted copy of the output boundary map $\bar{b}_\eta[n]$ onto the smoothed colors $\bar{\mathbf{f}}[n]$. 

\begin{supfigure}[H]
\begin{subfigure}{\textwidth}
\includegraphics[width=\columnwidth]{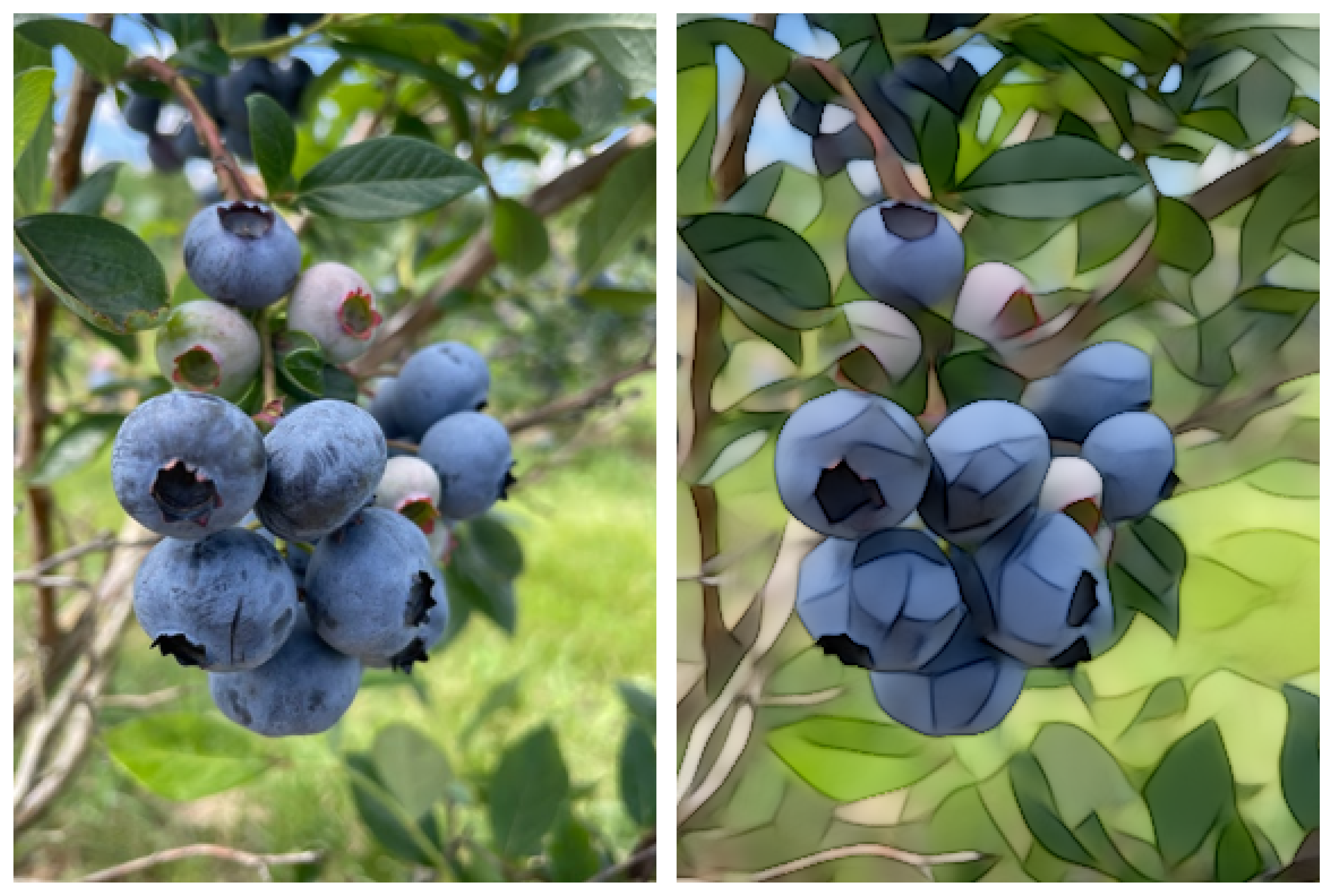}
\end{subfigure}
\begin{subfigure}{\textwidth}
\includegraphics[width=\columnwidth]{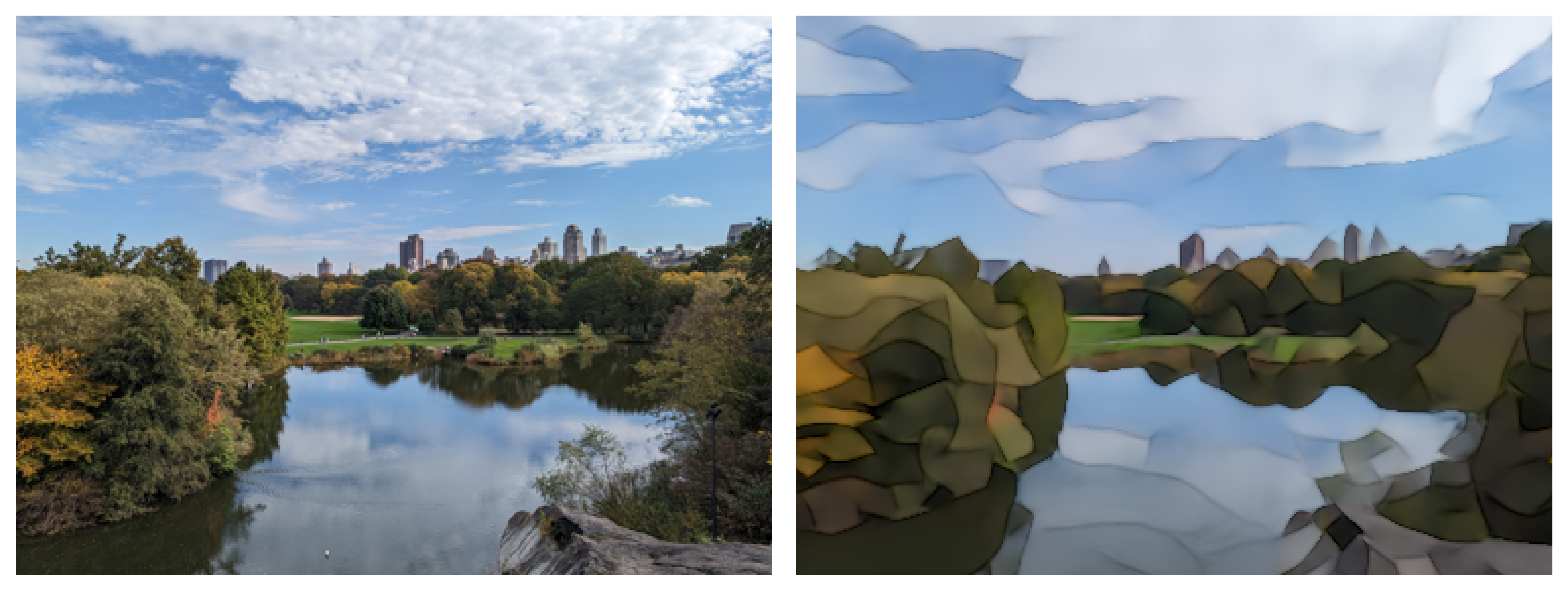}
\end{subfigure}
\caption{\label{fig:stylization} Examples of stylized natural photographs, created by imposing our method's output boundary map onto the output smoothed colors.}
\end{supfigure}

\newpage

\ifSubfilesClassLoaded{
\bibliographystyle{splncs04}
\bibliography{refs}
}{}

\end{document}

\bibliography{refs}

\end{document}